\documentclass{article}

\usepackage{microtype}
\usepackage{graphicx}
\usepackage{subcaption}
\usepackage{booktabs} %

\usepackage[accepted]{icml2025}

\usepackage{amsmath}
\usepackage{amssymb}
\usepackage{mathtools}
\usepackage{amsthm}

\usepackage{multirow}
\usepackage{hyperref}
\usepackage{xcolor}
\usepackage{xspace}
\usepackage{colortbl}
\usepackage{makecell}
\usepackage{bbding}
\usepackage{listings}
\usepackage{amssymb}
\usepackage{calc}
\usepackage{adjustbox}
\definecolor{customcolor}{HTML}{004496}
\definecolor{custompink}{HTML}{FF69B4}
\hypersetup{
colorlinks=true,
linkcolor=red,
citecolor=customcolor,
urlcolor=custompink
}
\usepackage[table]{xcolor}
\definecolor{rowgray}{gray}{0.95}
\usepackage{amssymb} %
\usepackage{pifont}  %
\usepackage{enumitem}

\newcommand{\cmark}{\ding{51}} %
\newcommand{\xmark}{\ding{55}} %

\usepackage[capitalize,noabbrev]{cleveref}

\theoremstyle{plain}

\theoremstyle{definition}

\theoremstyle{remark}

\newcommand{\x}{\mathbf{x}}
\newcommand{\T}{\mathcal{T}}
\newcommand{\bds}{\boldsymbol}
\newcommand{\dtrain}{\mathcal{D}_{\textrm{train}}}
\newcommand{\dtest}{\mathcal{D}_{\textrm{test}}}
\newcommand{\ours}{GenRe$^2$\xspace}

\icmltitlerunning{Beyond Token-level Supervision: Unlocking the Potential of Decoding-based Regression via Reinforcement Learning}

\begin{document}

\twocolumn[
\icmltitle{Beyond Token-level Supervision: Unlocking the Potential of \\ Decoding-based Regression via Reinforcement Learning}

\icmlsetsymbol{equal}{*}

\begin{icmlauthorlist}
\icmlauthor{Ming Chen}{equal,lamda,njuai}
\icmlauthor{Sheng Tang}{equal,lamda,njuai}
\icmlauthor{Rong-Xi Tan}{equal,lamda,njuai}
\icmlauthor{Ziniu Li}{cuhksz}
\icmlauthor{Jiacheng Chen}{cuhk}
\icmlauthor{Ke Xue$~^\dagger$}{lamda,njuai}
\icmlauthor{Chao Qian$~^\dagger$}{lamda,njuai}
\end{icmlauthorlist}

\icmlaffiliation{lamda}{National Key Laboratory for Novel Software Technology, Nanjing University}
\icmlaffiliation{njuai}{School of Artificial
Intelligence, Nanjing University}
\icmlaffiliation{cuhksz}{School of Data Science, The Chinese University of Hong Kong, Shenzhen}
\icmlaffiliation{cuhk}{Department of Computer Science and Engineering, The Chinese University of Hong Kong}

\icmlcorrespondingauthor{Ke Xue}{xuek@lamda.nju.edu.cn}
\icmlcorrespondingauthor{Chao Qian}{qianc@lamda.nju.edu.cn}

\icmlkeywords{Machine Learning, ICML}

\vskip 0.3in
]

\printAffiliationsAndNotice{\icmlEqualContribution} %

\begin{abstract}
Decoding-based regression, which reformulates regression as a sequence generation task, has emerged as a promising paradigm of applying large language models for numerical prediction. However, its progress is hindered by the misalignment between discrete token-level objectives (e.g., cross-entropy) and continuous numerical values. Existing approaches relying on token-level constraints often fail to capture the global magnitude of the target value, limiting their precision and generalization. In this paper, we propose to unlock the potential of decoding-based regression via Reinforcement Learning (RL). We formulate the generation process as a Markov Decision Process, utilizing sequence-level rewards to enforce global numerical coherence. Extensive experiments on tabular regression and code metric regression demonstrate that our method (specifically with ReMax and GRPO) consistently outperforms both state-of-the-art token-level baselines and traditional regression heads, showing the superiority of introducing sequence-level signals. Our analysis further reveals that RL significantly enhances sampling efficiency and predictive precision, establishing decoding-based regression as a robust and accurate paradigm for general-purpose numerical prediction.
\end{abstract}

\section{Introduction}
\label{submission}

Regression, the task of predicting continuous targets from input representations, stands as a fundamental role of machine learning~\citep{prml-book, position-tabular}, with wide applications across critical domains ranging from scientific discovery~\citep{hu2024reducing} to industrial scenarios~\citep{HE2025application}. Traditional regression methods, including Gaussian Processes~\cite{gpml} and tree-based models~\cite{xgboost, catboost}, excel due to their robustness and interpretability~\citep{sahakyan2021explainable}. However, with the advent of the deep learning era and the increasing complexity of data, there has been a paradigm shift towards deep-learning (DL) based regressors~\citep{dl-tabular-survey, talent-2}. These methods leverage the power of representation learning to map high-dimensional inputs into latent spaces, subsequently modeling the target value through specialized regression heads.

For DL-based regressors, there have been some design philosophies of regression heads to map latent representations to continuous targets. The most common approach, the pointwise head, projects representations directly to a scalar but often fails to capture the uncertainty or the complex multimodality of the target distribution~\citep{uncertainty-1}. To address this, parametric distribution heads model outputs as predefined distributions (e.g., Gaussian), yet they rely on rigid assumptions that may not hold in real-world scenarios~\cite{histogram-head}. Alternatively, the Riemann head (or histogram head) discretizes the continuous output into finite bins, converting regression into classification~\cite{hist-rl-1, histogram-head}, showing great robustness~\citep{histogram-head-2} and performance~\citep{pfn}.  However, these methods primarily operate on structured data, limiting their ability to perform regression on the vast and diverse spectrum of unstructured data (e.g., text or code). 

\begin{figure}[!t]
    \centering
\includegraphics[width=\linewidth]{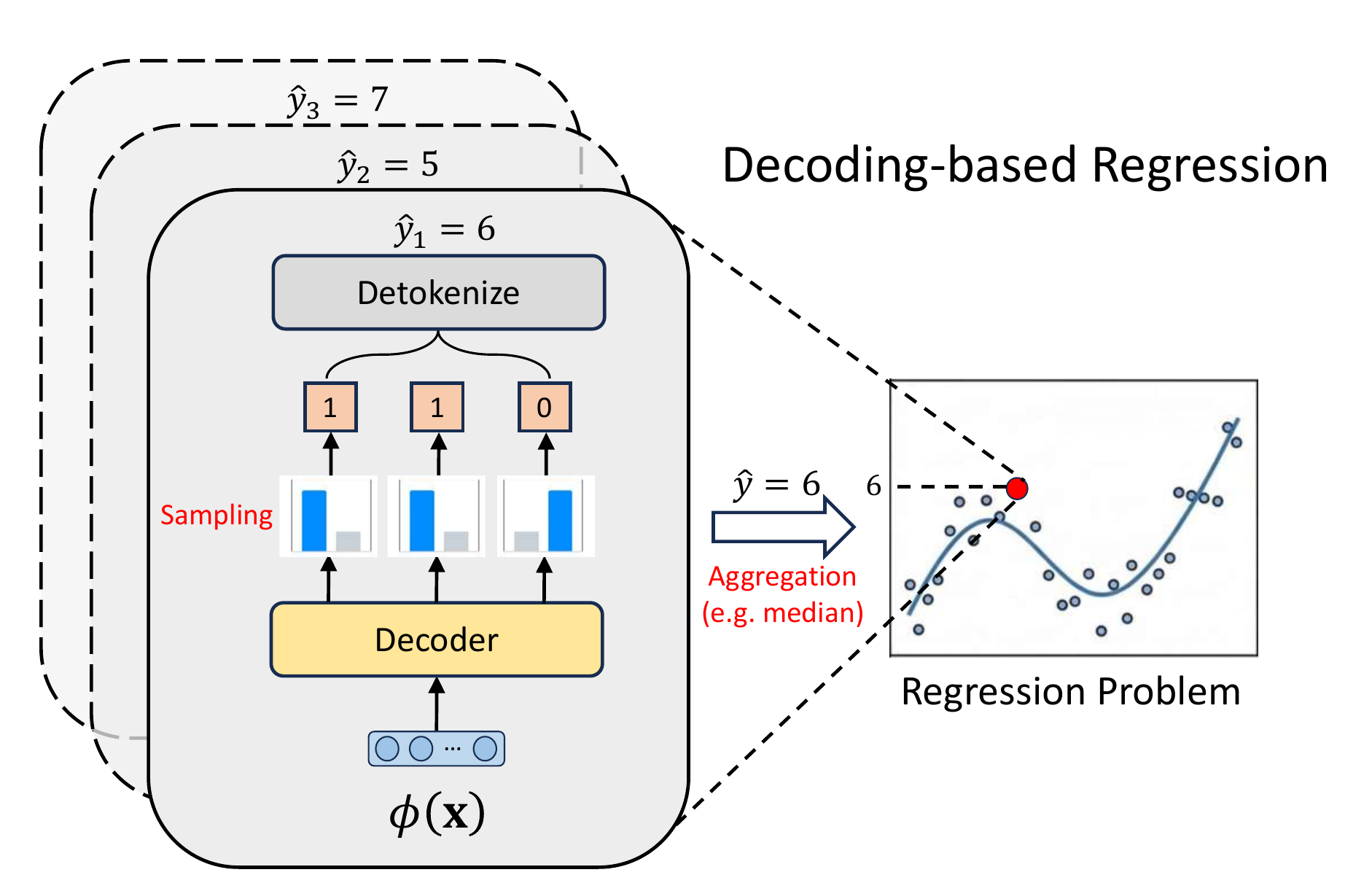}
    \vspace{-1.5em}
    \caption{Illustration of decoding-based regression. 
    The input $\mathbf{x}$ passes through an encoder to produce the representation $\phi(\mathbf{x})$, which is then processed by a decoder. The model performs multiple sampling trials to generate several discrete token sequences (e.g., the binary representation \texttt{<1><1><0>}). These sequences are individually detokenized into corresponding scalar values (shown in the stacked layers as $\hat{y}_1=6, \hat{y}_2=5, \hat{y}_3=7$). Finally, these scalar values are combined via an aggregation strategy (e.g., median) to produce the final prediction $\hat{y}=6$.}
    \label{fig:bg}
    \vspace{-1.5em}
\end{figure}

This limitation has motivated recent studies to leverage Large Language Models (LLMs) for universal regression~\citep{w2n, omnipred, RiR}. A key development in this line of work is decoding-based regression~\citep{decoding_regression}, which reformulates regression as a discrete sequence generation task and can be trained over large amounts of regression data $(\x,y)$ represented as text.
As illustrated in Figure~\ref{fig:bg}, this approach reformulates regression as a next-token prediction task by tokenizing continuous values (e.g., via base-$B$ expansion).
Unlike traditional scalar regressors, decoding-based regression not only can handle unstructured raw data, but also leverages the strong sequential modeling capabilities of Transformers to capture complex distributions~\citep{omnipred}.
Furthermore, the generative approach of decoding-based regression mitigates the susceptibility to reward hacking often seen in scalar or histogram baselines~\citep{reward-hacking-1,reward-hacking-2}, producing more robust and calibrated predictions, which align with the recent observations from  generative reward models~\cite{gen-rm,gen-verifier}.
The concept of decoding-based regression gives rise to Regression Language Model (RLM)~\citep{omnipred}, which demonstrates great potential in diverse applications ranging from industrial prediction~\citep{regress_lm,code_rlm} to black-box optimization~\citep{embed-then-regress,uniso}.

However, despite its promise, the potential of decoding-based regression remains unlocked. The critical barrier lies in the misalignment between the widely used Cross-Entropy (CE) loss and the numerical nature of the regression task~\citep{RAFT-1}. CE treats tokens as independent categories, ignoring their ordinal value and the entire magnitude of the detokenized number. While recent works have attempted to mitigate this via token-level distance penalties, e.g., NTL~\cite{ntl} and DIST$^2$~\citep{DIST2}, a fundamental limitation remains: these methods operate locally on individual tokens and overlook the cumulative error over the entire sequence~\citep{LLMReg-fail-precise}, which can lead to catastrophic outcomes in the original numerical space~\citep{omnipred, decoding_regression}. Thus, there is an urgent need for a method that is inherently aware of sequence-level numerical magnitude.

In this paper, we propose \textbf{Gen}erative \textbf{Re}inforced \textbf{Re}gressor (GenRe$^2$) to bridge this gap. We reformulate decoding-based regression as a Markov Decision Process (MDP), allowing us to optimize the model using policy gradient methods~\citep{pg-method}. Unlike previous approaches, \ours utilizes a sequence-level reward signal, which is computed only after the full numerical sequence is generated and detokenized, to directly guide the model towards minimizing the true regression error (e.g., MSE). We explore efficient REINFORCE-style~\citep{reinforce} algorithms, including ReMax~\cite{remax} and GRPO~\cite{GRPO} to finetune the CE-trained models, and validate \ours across two distinct domains: tabular regression on the TALENT benchmark~\cite{talent-1,talent-2} and code metric regression~\citep{code_rlm} using RLM~\citep{omnipred, regress_lm, code_rlm}. Experimental results demonstrate that \ours consistently outperforms the pointwise and Riemann baselines, and state-of-the-art token-level improvements for decoding-based regressor, clearly showing the benefits of \ours based on sequence-level reward.

Our findings reveal that (1) Equipped with \ours, the decoding-based paradigm generally outperforms traditional designs (e.g., pointwise and Riemann heads); (2) Sequence-level supervision is significant for decoding-based regression to bridge the gap between regression and the token-level objectives; (3) While RL may sharpen the output distribution, it significantly enhances the sampling efficiency and precision, making generative decoding-based models as a highly competitive paradigm for numerical prediction.

\section{Background}
\label{sec:background}

Traditional DL-based regressors typically employ a pointwise head (predicting a scalar) or a Riemann head (predicting binned histogram distribution)~\citep{histogram-head}, where the Riemann head has better robustness and performance in many applications and is widely used~\citep{hist-rl-2,tabpfn}. A detailed overview of these methods is provided in Appendix~\ref{sec: reg}. Recently, \citet{decoding_regression} proposed decoding-based regression by reformulating regression as a discrete sequence generation task, calling for a paradigm shift to generative regression. 
Specifically, a target scalar value $y$ is transformed into a sequence of discrete tokens $\mathcal{T}=\{t_1, t_2,\cdots, t_K\}$. Then, an autoregressive decoder head is trained to predict the tokens sequentially. Given the input representation $\phi(\x)$, it models the conditional probability distribution $p_{\bds{\theta}}(y|\x)$ as
\begin{align*}
    p_{\bds{\theta}}(y|\x) = \prod\nolimits_{k=1}^K p_{\bds{\theta}}(t_k\mid\phi(\x),\mathcal{T}_{<k}),
\end{align*}
where $\mathcal{T}_{<k}$ denotes the tokens generated before step $k$. Given a dataset $\mathcal{D}=\{(\x_i, y_i)\}_{i=1}^N$, we first tokenize each target $y$ into a corresponding token sequence $\T$, then the decoder head is trained to predict the next token by minimizing the standard Cross-Entropy (CE) loss: 
\begin{align*}
    \mathcal{L}(\bds{\theta})=-\mathbb{E}_{(\x, \T)\sim\mathcal{D}}\left[\sum\nolimits_{k=1}^K \log p_{\bds{\theta}}\left(t_{k}\mid\phi(\x),\mathcal{T}_{<k}\right)\right].
\end{align*}

For inference, we generate $m$ candidate solutions via sampling (e.g., temperature sampling) and return the aggregation of these solutions.
Here, the aggregation strategy can be various, such as $\operatorname{mean}(\cdot)$ or $\operatorname{median}(\cdot)$, and different aggregations lead to Bayes-optimal solutions for different regression metrics~\citep{RAIL}. 

The tokenization of decoding-based regression is important. Following~\citep{decoding_regression}, we briefly introduce two common tokenization strategies (Detailed description can be founded in Appendix~\ref{sec:tokenization}):
\vspace{-0.5em}
\begin{enumerate}[leftmargin=1em, labelindent=0em]
    \item [$\bullet$] \textit{Normalized Tokenization}: 
    The normalization tokenization first scales a target value $y$ to a fixed interval (e.g. $[0, 1]$), then represents the scaled value as a base-$B$ expansion (e.g., $0.6$ as \texttt{<1><1><0>} with $B=2$).  
    While effective, it relies on the access to the global minimum and maximum and is highly sensitive to outliers~\citep{power-transform, omnipred}. 
    \vspace{-0.25em}
    \item[$\bullet$] \textit{Scientific Notation Tokenization}:
    Scientific notation tokenization methods (e.g., P10~\citep{P10} or IEEE~\citep{IEEE}) do not normalize the target, representing numbers using sign, mantissa, and exponent components (e.g., P10 represents $1.23$ as \texttt{<+><1><2><3><E-2>}). 
    This tokenization supports a wider range of values but can be prone to yield hallucinations in unbounded generation~\citep{omnipred}. 
\end{enumerate}
\vspace{-0.5em}

Intuitively, decoding-based regression generalizes histogram-based regression (e.g., Riemann head) into a multi-step binning paradigm, where tokenization defines the structure and the autoregressive decoding sequentially refines predictions~\citep{decoding_regression}.
Notably, this approach offers two clear advantages: 
(1) It integrates seamlessly with LLMs, thereby enabling universal regression~\citep{omnipred} on free-formed inputs~\citep{regress_lm} while leveraging rich priors~\citep{code_rlm, machinelearninglm}; 
(2) It improves calibration. As noted in the reward model community~\citep{GenRM-2, GenRM-3}, sequential generative scoring yields more robust predictions~\citep{RAFT-2} and better mitigates reward hacking compared to scalar or histogram baselines~\citep{reward-hacking-1, reward-hacking-2}.

Decoding-based regression has been applied to many downstream scenarios~\citep{physix, capellm}, one representative of which is Regression Language Model (RLM)~\citep{omnipred}. 
RLM directly regresses in the form of natural language, eliminating the need of feature engineering, which has been successfully applied to industrial scenarios~\citep{regress_lm}, code metric prediction~\citep{code_rlm}, and black-box optimization~\citep{embed-then-regress,uniso}.

\section{Method}
\label{sec:method}
In this section, we present our proposed method, \ours, which leverages RL to address the sequence-level challenge of decoding-based regression with policy gradient. 
In Section~\ref{sec:limitation}, we first discuss the limitations of previous token-level decoding-based regression methods, showing emergent need for sequence-level supervisions and motivating us to solve it via RL.
In Section~\ref{sec:formulation}, we formulate the decoding-based regression task as a Markov Decision Process (MDP)~\citep{MDP-book}, which serves as the foundation of \ours.
In Section~\ref{sec:reward_engineering}, we discuss several reward design strategy to guide the model towards better regression performance. 
Finally, we present and visualize some training dynamics of \ours in Section~\ref{sec:training_dynamics}.

\subsection{Limitations of Previous Token-level Methods}
\label{sec:limitation}
\begin{figure}[!t]
    \centering
    \includegraphics[width=\linewidth]{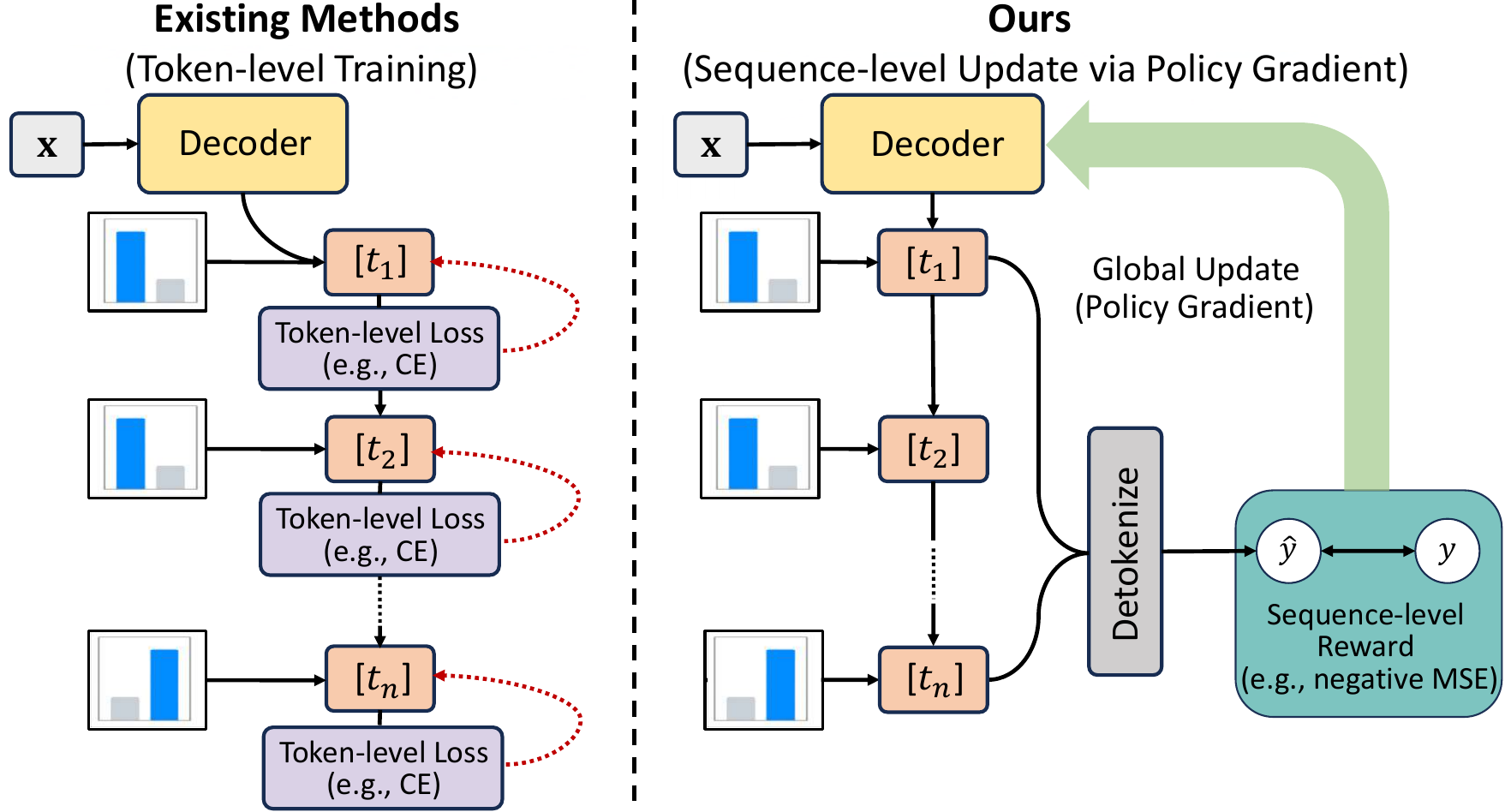}
    \vspace{-1.5em}
    \caption{Comparison between local token-level training and global sequence-level update.  
    \textbf{Left} (existing methods): The model is trained at each token $[t_1, \dots, t_n]$ with a local loss (e.g., CE) that focuses solely on individual tokens. 
    \textbf{Right} (ours): The model generates a full sequence and detokenizes it into a prediction $\hat{y}$. A global reward (i.e., negative MSE) against the ground truth $y$ is then backpropagated to update the model parameters.
    }
    \label{fig:token-level-cons}
    \vspace{-0.5em}
\end{figure}

Standard decoding-based regression typically relies on CE.
However, the potential of CE-trained decoding-based regression remains locked.
\citet{RAFT-1, LLMReg-fail-precise} theoretically showed that CE is not well-aligned with regression, as it treats digits as individual categories and ignores the numerical continuity. 
While recent improvements like NTL~\citep{ntl} and DIST${}^2$~\citep{DIST2} introduce distance penalties, they still operate locally on individual tokens.
As illustrated in the left part of Figure~\ref{fig:token-level-cons}, the token-level losses overlook the global magnitude of the detokenized number.
However, the true regression error is determined by the holistic value of the generated sequence, showing a misalignment with the token-level supervisions.

To bridge this gap, we reformulate the task via RL, optimizing the decoder using policy gradients with full-sequence rewards (Figure~\ref{fig:token-level-cons}, right).
This approach is motivated by recent successes in RL for LLM~\citep{llm-rl-survey-1}, where policy gradient methods~\citep{policy-gradient,pg-method} effectively align LLMs' responses with sequence-level, non-differentiable objectives, such as human preference~\citep{RLHF-0, RLHF-1} or verifiable correctness~\citep{deepseek-r1, llm-math-survey}. 
We provide a detailed overview of RL for LLMs in Appendix~\ref{appendix:RL4LLM}.
Next, we will introduce the RL formulation of decoding-based regression, establishing the foundation of \ours.

\subsection{Problem Formulation}
\label{sec:formulation}
In this section, we formalize the generation of a numerical sequence (i.e., the primary goal of decoding-based regression) as an MDP.
Specifically, taking the generation of $6$ (the sequence representation is \texttt{<1><1><0>}) as an example, for an input representation $\phi(\mathbf{x})$, the MDP $\mathcal{M}=(\mathcal{S}, \mathcal{A}, P, r, \rho, T)$ can be written by:
\begin{itemize}[leftmargin=1em, labelindent=0em]
   \item \textbf{State} $\mathcal{S}$: A state $s_k \in \mathcal{S}$ is defined by the input feature and the
    generated token sequence, i.e., $s_k = (\phi(\mathbf{x}), \T_{<k})$, where $\mathcal{T}_{<k} = (t_0, \dots, t_{k-1})$. For instance, an intermediate state at $k=2$ is $s_2=(\phi(\x), \texttt{<1><1>})$.
    \item \textbf{Action} $\mathcal{A}$: The action space $\mathcal{A}$ is defined over the token vocabulary $\mathcal{V}$, where
    an action $a_k$ is the selection of the next token $t_k \in \mathcal{V}$. For instance, given the state $s_2$, the model can sample the next token \texttt{<0>} or \texttt{<1>} to proceed towards completing the sequence.
    \item \textbf{Transition} $P$: The state transitions $P(s_{k+1}\mid s_k, a)$ are deterministic. 
    Appending a selected token $a_k=t_k$ to the current state $s_k$ always leads to a unique next state $s_{k+1} = (s_k,a_k)= (\phi(\mathbf{x}), \{\T_{<k}, t_k\})$, which is also an important characteristic of RL formulation in LLM~\citep{remax, llm-rl-survey-1}. Continuing the example from state $s_2$, if the model samples the action \texttt{<0>}, the state transitions to a sequence \texttt{<1><1><0>} (decoding to 6); conversely, if the action \texttt{<1>} is sampled, the state updates to \texttt{<1><1><1>} (decoding to 7).
    \item \textbf{Reward} $r$: The reward function $r$ assigns reward values to state-action pairs. 
    Since we have to access signals from the detokenized numerical value only after the entire sequence is generated, the reward function is defined by:
    \begin{align*}
        r(s_k,a_k)=\begin{cases}
            0 & \text{if } k\neq K-1 \\
            r(\phi(\x), a_{0:K-1})&\text{otherwise.}
        \end{cases}
    \end{align*}
    Consistent with the formulation of RL in LLM~\citep{llm-rl-survey-1, llm-rl-survey-2}, this reward design is sparse with zero rewards to all intermediate generation steps.
    The specific design of $r(\phi(\x), a_{0:K-1})$ can be flexible, which we will elaborate in Section~\ref{sec:reward_engineering}. 
    \item \textbf{Initial State Distribution} $\rho$: The distribution $\rho$ is deterministic, with the initial state $s_0$
    corresponding to the input feature $\phi(\mathbf{x})$ and an empty sequence.
    \item \textbf{Horizon} $T$: $T$ is the maximum length of the generated sequence, i.e., $T=K$.
\end{itemize}
Within this framework, the learning objective of \ours is to maximize the expected return:
\begin{align}\label{eq:objective}
    \mathcal{J}(\pi_{\bds{\theta}}) = \mathbb{E}_{(\x, y) \sim \dtrain }\mathbb{E}_{\tau\sim\pi_{\bds\theta}} \left[ \sum\nolimits_{k=0}^{K-1} r(s_k,a_k) \right],
\end{align}
where $\pi_{\bds\theta}$ is the policy parameterized by $\bds\theta$, and $\tau=(s_0, a_0,\cdots,s_K)$ denotes a trajectory sampled from $\pi_{\bds\theta}$. 

By formulating the decoding regression task into a policy optimization problem, we employ the policy gradient method~\citep{pg-method} to optimize $\pi_{\bds\theta}$ by ascending the gradient of the expected return:
\begin{equation}\label{eq:pg-estimate}
\begin{aligned}
   & \nabla_{\bds\theta}\mathcal{J}(\pi_{\bds\theta})=\mathbb{E}_{(\x, y) \sim \dtrain }\mathbb{E}_{\tau\sim\pi_{\bds\theta}} \\
    &\qquad \left[ \sum\nolimits_{k=0}^{K-1} \nabla_{\bds\theta} \log \pi_{\bds\theta}(a_k\mid s_k)A^{\pi_{\bds\theta}}(s_k, a_k) \right],
\end{aligned}
\end{equation}
where $A^{\pi_{\bds\theta}} (s_t, a_t)$ is the advantage function estimating the relative value of action $a_k$ in state $s_k$. Given the deterministic state transitions in the MDP, simple policy gradient methods like REINFORCE~\citep{reinforce} are efficient. In this work, we employ two prevalent REINFORCE-style algorithms, ReMax~\citep{remax} and GRPO~\citep{GRPO}, details of which are provided in Appendix~\ref{sec:pg_methods}.

\subsection{Reward Design for \ours}
\label{sec:reward_engineering}
The reward function is designed to guide the model towards the final regression metrics. 
Upon completing an episode, the generated sequence $\tau$ is detokenized into its original prediction $\hat{y} = \operatorname{Detokenize}(\tau)$. 
Given a bijective mapping $\psi$, we can define the terminal reward via the distance-based metrics in the target space, e.g., via the negative Mean Squared Error (MSE):
\begin{equation}\label{eq:reward}
    R(\tau) = -(\psi(\hat{y}) - \psi(y))^2 ,
\end{equation}
where $y$ is the ground-truth target~\footnote{Given the inherent quantization error by discrete tokenization~\citep{LLMReg-fail-precise}, one could round the target $y$ to the nearest tokenization bin to calculate the metrics. However, we omit this detail for simplicity.}. 
The mapping $\psi$ can be chosen flexibly, e.g., identity or normalization.
This reward function is calculated on sequence level, and thus inherently numerically aware on the target space, which is a property being ignored in previous token-level objectives~\citep{ntl, DIST2}. 
It correctly assigns a relative higher reward to a numerically close prediction (e.g., \texttt{101} for a target of \texttt{100}) compared to a numerically distant one (e.g., \texttt{200}), even if both differ by a single token. 
This directly forces the model to learn the principles of numerical magnitude and proximity. 
We will discuss different settings of $\psi$ according to different problem natures in Section~\ref{sec:exp}.

Notably, compared to Reinforcement Learning with Verifiable Reward (RLVR) research in LLM~\citep{llm-rl-survey-1, llm-rl-survey-2}, which takes sparse reward, e.g., $\{-1, +1\}$, the reward in \ours is a dense one, where different generated sequences receive different rewards.

\subsection{Training Dynamics}
\label{sec:training_dynamics}
We follow the settings of~\citep{decoding_regression} to examine the feasibility of \ours.
Specifically, we instantiate the encoder $\phi$ as a Multi-Layer Perception (MLP), and the autoregressive decoder as a standard Transformer decoder with normalized tokenization. 
Here we use the negative MSE on the normalized space as the reward.
The RL training pipeline is implemented under the \texttt{accelerate} framework~\citep{accelerate} with \texttt{deepspeed}~\citep{deepspeed} ZeRO stage 2~\citep{zero}.
Analogous to the common practice of performing RL after SFT in LLM post-training, we initiate RL using the CE-trained checkpoint that achieved the minimum validation loss.

\begin{figure}
    \centering
    \includegraphics[width=0.49\linewidth]{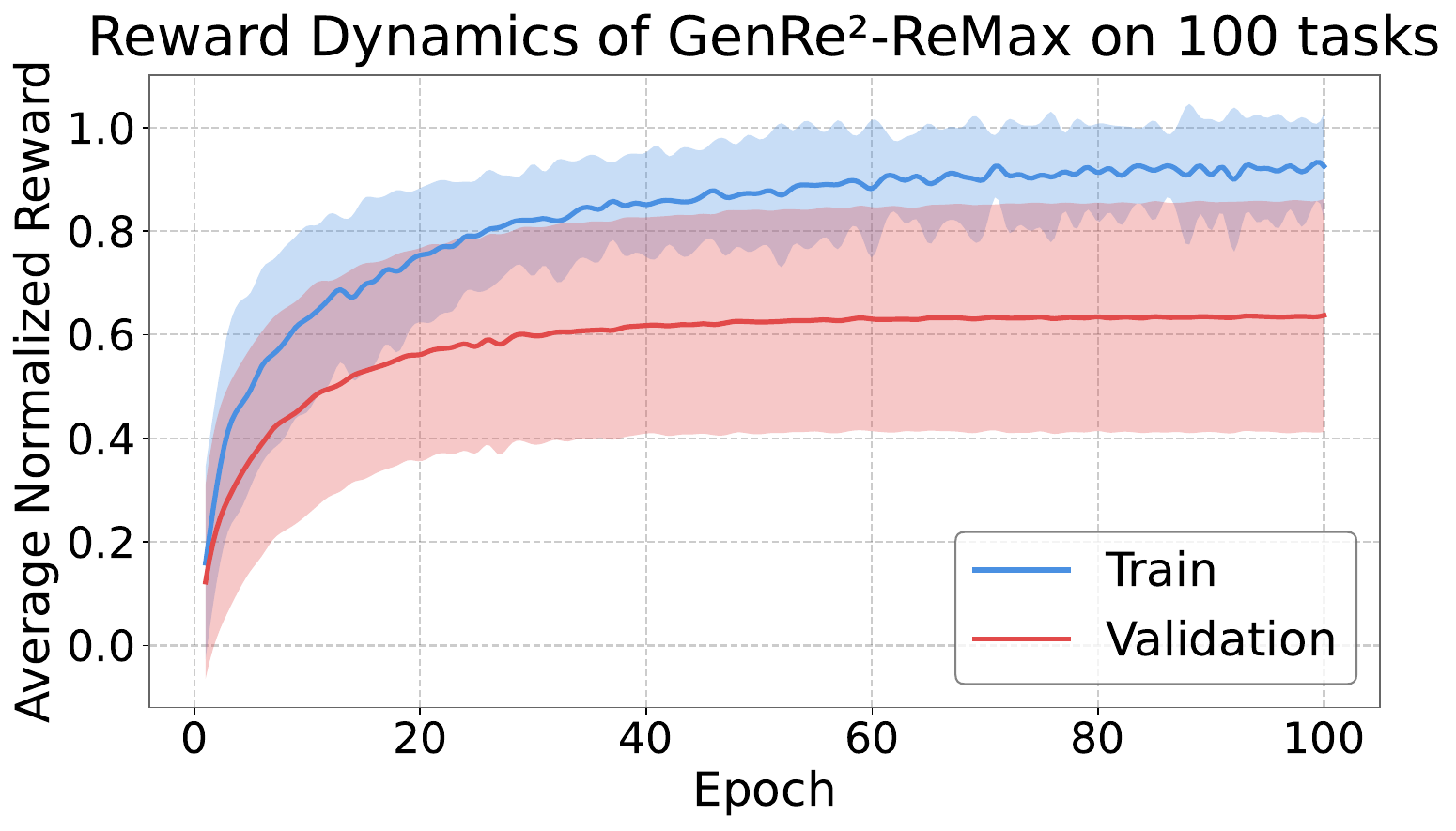}
    \includegraphics[width=0.49\linewidth]{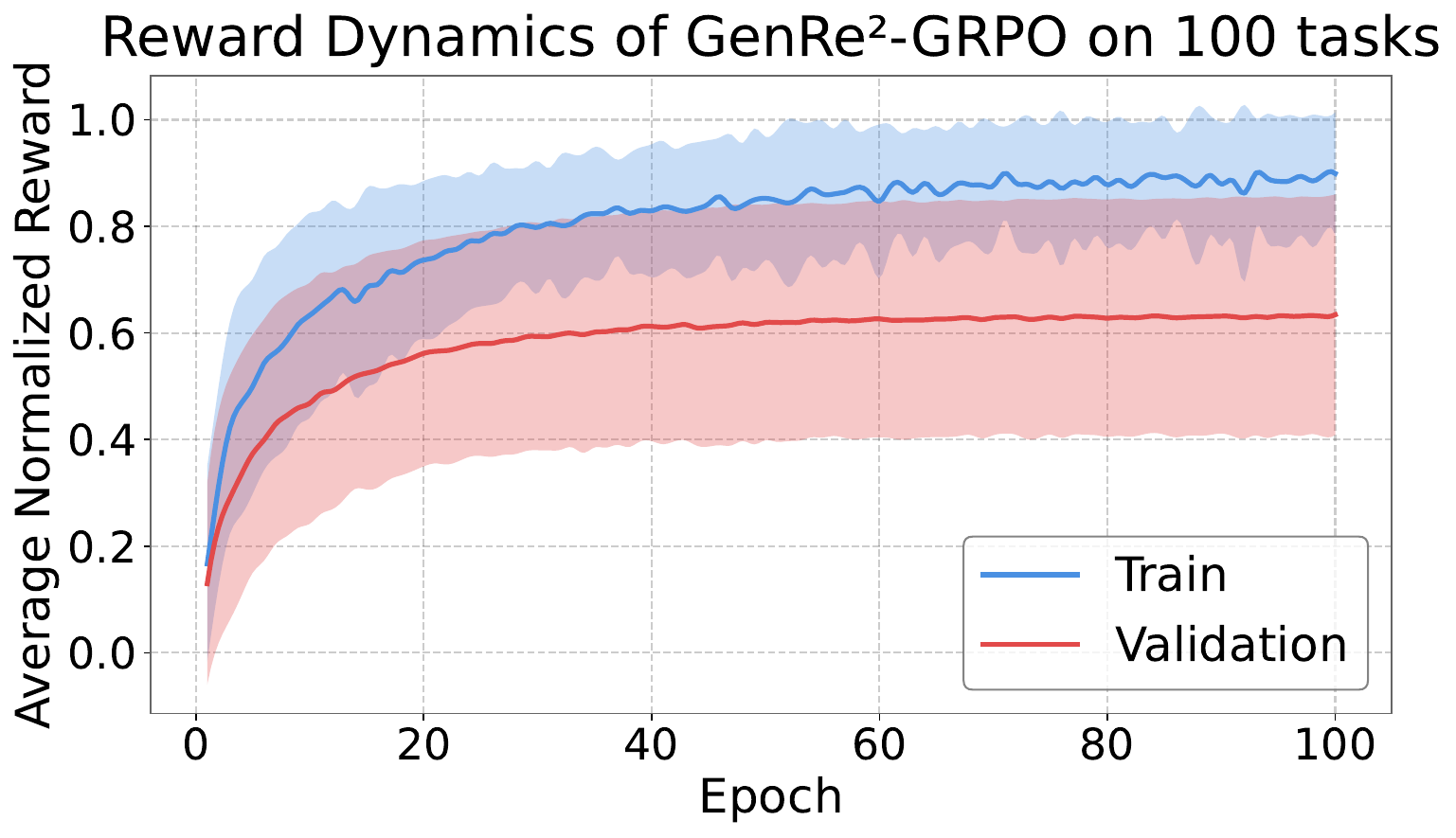} \\
    \includegraphics[width=\linewidth]{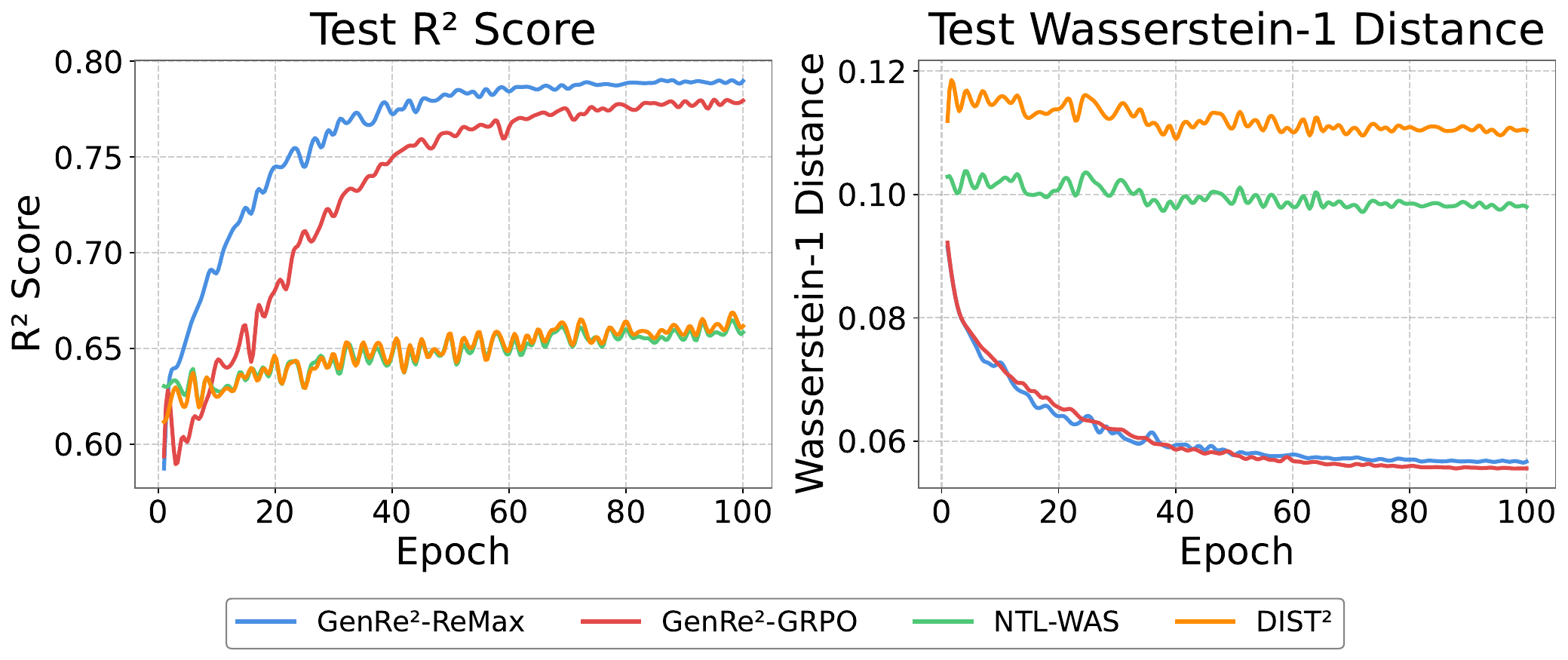}\vspace{-0.5em}
    \caption{Training dynamics of \ours. 
    \textbf{Top row}: Normalized reward dynamics for \ours combined with ReMax (left) and GRPO (right) on 100 TALENT regression tasks, where the reward is normalized to $[0,1]$ with respect to each task.  
    \textbf{Bottom row}: Visualization of regression performance dynamics on \texttt{Kaggle\_\allowdisplaybreaks bike\_\allowdisplaybreaks sharing\_\allowdisplaybreaks demand\_\allowdisplaybreaks challange}~\citep{kaggle}, comparing \ours with NTL-WAS~\citep{ntl} and DIST${}^2$~\citep{DIST2} on test R${}^2$ score (left, higher is better) and test Wasserstein-1 distance (right, lower is better).}
    \vspace{-1em}
    \label{fig:reward}
\end{figure}

\textbf{Reward dynamics.}
We run \ours on the TALENT benchmark~\citep{talent-1, talent-2}, expanding over 100 regression tasks. 
In the two top sub-figures of Figure~\ref{fig:reward}, we present the training and validation reward dynamics of \ours combined with ReMax and GRPO, where the rewards of individual tasks are normalized to $[0,1]$. 
It can be observed that the rewards increase steadily and result in stable convergence, showing that our method is robust across diverse tasks.

\textbf{Regression performance dynamics.}
We further analyze the regression performance dynamics of \ours on a representative dataset, \texttt{Kaggle\_\allowdisplaybreaks bike\_\allowdisplaybreaks sharing\_\allowdisplaybreaks demand\_\allowdisplaybreaks challange}~\citep{kaggle}.
Rather than focusing only on the final metrics, e.g., the coefficient of determination (R${}^2$), we consider the Wasserstein-1 distance to measure the distance between the output distribution (i.e., the histogram distribution of the generated candidates) and the target.
Formally, assume the output distribution $P$ lies in a group of supports $\{z_i\}_{i=1}^k$ with probabilities $\{p_i\}_{i=1}^k$, then we can calculate the Wasserstein-1 distance as:
$W_1=\sum_{i=1}^k p_i\cdot|z_i-y_{\textrm{true}}|$,
where $y_{\textrm{true}}$ represents the ground-truth target. The Wasserstein-1 distance quantifies the distance of $P$ towards $y_{\textrm{true}}$, where lower distance indicates better regression performance.
As shown in the two bottom sub-figures in Figure~\ref{fig:reward}, compared to other token-level methods, \ours achieves both significantly lower $W_1$ distance and better performance, demonstrating better alignment with the aim of regression.
This clearly shows the advantage of focusing on the global structure and numerical magnitude on sequence-level.

\section{Experiments}
\label{sec:exp}
In this section, we empirically compare \ours with a variety of baseline methods on two representative decoding-based regression tasks. 
In Section~\ref{sec:tabular}, we evaluate \ours on tabular regression tasks, while in Section~\ref{sec:code}, we conduct experiments on the recently proposed Regression Language Model (RLM) to address code-to-metric regression.
We show the performance of different methods and conduct case studies to analyze algorithmic behavior. 
Finally, we deliver some empirical discussions to understand the superior performance of RL in Section~\ref{sec:analysis}.

\begin{table*}[ht!]
\vspace{-1.0em}
\caption{RMSE, R${}^2$ and Rank Correlation results over 5 random seeds on 100 TALENT regression tasks. 
The best and runner-up results are \textbf{bolded} and \underline{underlined}, respectively.  
For the decoding-based methods, ``Median" and ``Mean" denote the aggregation strategy used to derive the final prediction from the generated candidates. 
Rows shaded in \colorbox{gray!15}{gray} indicate our proposed methods.}
\centering
\resizebox{1.0\linewidth}{!}{
\newcommand{\alignmytext}[1]{\makebox[\widthof{+NTL-WAS~\citep{ntl} }][l]{#1}}

\begin{tabular}{cc|cc|cc|cc}
    \toprule
    \multirow{2}{*}{Head} & \multirow{2}{*}{Method} & \multicolumn{2}{c|}{RMSE $\downarrow$} & \multicolumn{2}{c|}{R$^2$ $\uparrow$} & \multicolumn{2}{c}{Rank Corr. $\uparrow$} \\
     & & Median & Mean & Median & Mean & Median & Mean \\
    \midrule
    
    Pointwise & / 
    & \multicolumn{2}{c|}{0.5563$_{\pm0.0035}$}
    & \multicolumn{2}{c|}{0.5708$_{\pm0.0262}$}
    & \multicolumn{2}{c}{0.7289$_{\pm0.0046}$}\\ 
    \midrule
    
    Riemann & / 
    & \multicolumn{2}{c|}{0.5435$_{\pm0.0004}$}
    & \multicolumn{2}{c|}{0.6170$_{\pm0.0008}$}
    & \multicolumn{2}{c}{0.7709$_{\pm0.0006}$} \\
    \midrule

    & \alignmytext{Base Model} 
    & 0.5484$_{\pm0.0004}$ & 0.5327$_{\pm0.0004}$ 
    & 0.6124$_{\pm0.0004}$ & 0.6368$_{\pm0.0005}$ 
    & 0.7705$_{\pm0.0007}$ & 0.7670$_{\pm0.0011}$ \\ 
    
    & \alignmytext{+NTL-WAS~\citep{ntl}} 
    & 0.5478$_{\pm0.0006}$ & 0.5307$_{\pm0.0007}$ 
    & 0.6132$_{\pm0.0009}$ & \underline{0.6391$_{\pm0.0009}$} 
    & 0.7712$_{\pm0.0013}$ & 0.7689$_{\pm0.0003}$ \\ 
    
    & \alignmytext{+NTL-MSE~\citep{ntl}} 
    & 0.5478$_{\pm0.0012}$ & 0.5320$_{\pm0.0013}$ 
    & 0.6098$_{\pm0.0044}$ & 0.6343$_{\pm0.0049}$ 
    & 0.7721$_{\pm0.0011}$ & 0.7686$_{\pm0.0003}$ \\ 
    
    & \alignmytext{+DIST$^2$~\citep{DIST2}} 
    & 0.5457$_{\pm0.0019}$ & 0.5810$_{\pm0.0017}$ 
    & 0.6096$_{\pm0.0057}$ & 0.4678$_{\pm0.0090}$ 
    & \underline{0.7734$_{\pm0.0007}$} & 0.7334$_{\pm0.0018}$ \\  
    
    \cmidrule{2-8}
    
    \rowcolor{rowgray} \cellcolor{white}
    & \alignmytext{\textbf{+\ours-ReMax (Ours)}} 
    & \textbf{0.5190$_{\pm0.0014}$} & \textbf{0.5151$_{\pm0.0012}$} 
    & \textbf{0.6459$_{\pm0.0020}$} & \textbf{0.6508$_{\pm0.0017}$} 
    & \textbf{0.7785$_{\pm0.0011}$} & \underline{0.7728$_{\pm0.0017}$} \\ 
    
    \rowcolor{rowgray} \cellcolor{white}\multirow{-7}{*}{Decoder}
    & \alignmytext{\textbf{+\ours-GRPO (Ours)}} 
    & \underline{0.5320$_{\pm0.0020}$} & \underline{0.5271$_{\pm0.0019}$} 
    & \underline{0.6248$_{\pm0.0062}$} & 0.6316$_{\pm0.0060}$ 
    & \textbf{0.7785$_{\pm0.0011}$} & \textbf{0.7737$_{\pm0.0016}$} \\
    \bottomrule
\end{tabular}
}
\vspace{-1em}
\label{tab:tabular-res-combined}
\end{table*}

\subsection{Tabular Regression}
\label{sec:tabular}
We examine the ability of \ours to perform tabular regression on TALENT benchmark~\citep{talent-1, talent-2}, a popular benchmark for tabular data containing 100 regression datasets. 
Following the practice of conducting RL after SFT in LLM post-training, we start RL from the CE-pretrained checkpoints. In Section~\ref{sec:tabular-setttings}, we introduce our experimental settings. 
Then we present the results to show the superiority of \ours in Section~\ref{sec:tab-results}. 
We also examine the robustness of \ours across multiple tokenization settings, and give explanations for the different performance when combined with ReMax and GRPO.
\subsubsection{Experimental Setup}
\label{sec:tabular-setttings}

\textbf{Compared methods.} 
We mainly consider two categories of methods: (1) Baselines with different regression heads (i.e., pointwise head and Riemann head); (2) Decoding-based regression methods, including two NTL variants (NTL-WAS and NTL-MSE)~\citep{ntl} and DIST$^2$~\citep{DIST2}.
Here, NTL and DIST$^2$ are improvement methods for decoding-based regression with token-level loss.
Details of the compared baselines can be found in Appendix~\ref{sec:baseline-details}. 
Following~\citep{decoding_regression}, we instantiate the encoder $\phi$ as an MLP, and the decoder is a standard Transformer decoder.

\textbf{Implementation details.}
Following the common protocol in tabular research~\citep{talent-1, talent-2}, we standardize the input $\x$ using z-score transformation.
We train the pointwise baseline with MSE loss, and the Riemann baseline with the unbounded variants suggested by~\citep{pfn, pfns4bo}. 
For other decoding-based baselines and \ours, we first train the base model from scratch using CE loss for 200 epochs, followed by fine-tuning the best validation checkpoint for 100 epochs using the respective strategies.
We use the normalized tokenization to prevent outliers~\citep{decoding_regression} by default, with the digit base $B=2$ and output sequence length $K=8$. 
Other details, including objective normalization and model optimization, can be found in Appendix~\ref{appendix:norm_head} and \ref{sec:exp_detail}. 
We will also consider different tokenization settings in Section~\ref{sec:tab-results}.

\textbf{RL details.}
We set the rollout budget $G=16$ in the experiments. 
The reward function is the negative MSE in Eq.~\eqref{eq:reward}.
Before calculating the reward, we first transform the detokenized number to its original space, and then set the mapping $\psi$ as a z-score standardization $\psi(y)=\frac{y-\mu_y}{\sigma_y}$ where $\mu_y$ and $\sigma_y$ represent the mean and standard derivation of $y$ in the training set, respectively.
We perform this transformation for fair comparison to the pointwise and Riemann baselines, which also conduct z-score on target scores.
We employ two RL methods (i.e., ReMax~\citep{remax} and GRPO~\citep{GRPO}) to finetune the pretrained checkpoint using AdamW optimizer~\citep{adamw} with an initial learning rate of $5\times 10^{-5}$, and report results of the checkpoint that achieves the best validation reward.

\textbf{Evaluation.}
We compare all methods on a suite of regression metrics, including RMSE, R$^2$, and Spearman's Rank Correlation. 
For decoding-based methods, following~\citep{regress_lm}, we directly sample from the model's output distribution with temperature 1.0 to generate $m=128$ candidate solutions, and aggregate them via both mean and median.

\subsubsection{Results and Analyses}
\label{sec:tab-results}
\textbf{Main results.}
In Table~\ref{tab:tabular-res-combined}, we report the main results of our tabular regression experiments, where our method \ours is appended with the name of the employed RL backbone. 
We can observe that: 
(1) The base model of the decoder head method consistently outperforms the pointwise baseline, showing competitive performance against the Riemann baseline;
(2) All of the token-level methods, NTL-WAS, NTL-MSE, and DIST$^2$ do not consistently improve the performance of the base model after finetuning, with slight improvement on some metrics; 
(3) Our proposed methods, \ours-ReMax and \ours-GRPO, instead significantly improve the performance of the base model, where \ours-ReMax achieves the best overall performance on all metrics, and \ours-GRPO performs best on rank correlation and is runner-up on RMSE.
After finetuned by our methods, the decoding-based methods consistently outperform the pointwise and Riemann baselines, demonstrating superiority for regression modeling.

\begin{figure}[t!]
    \centering
    \includegraphics[width=\linewidth]{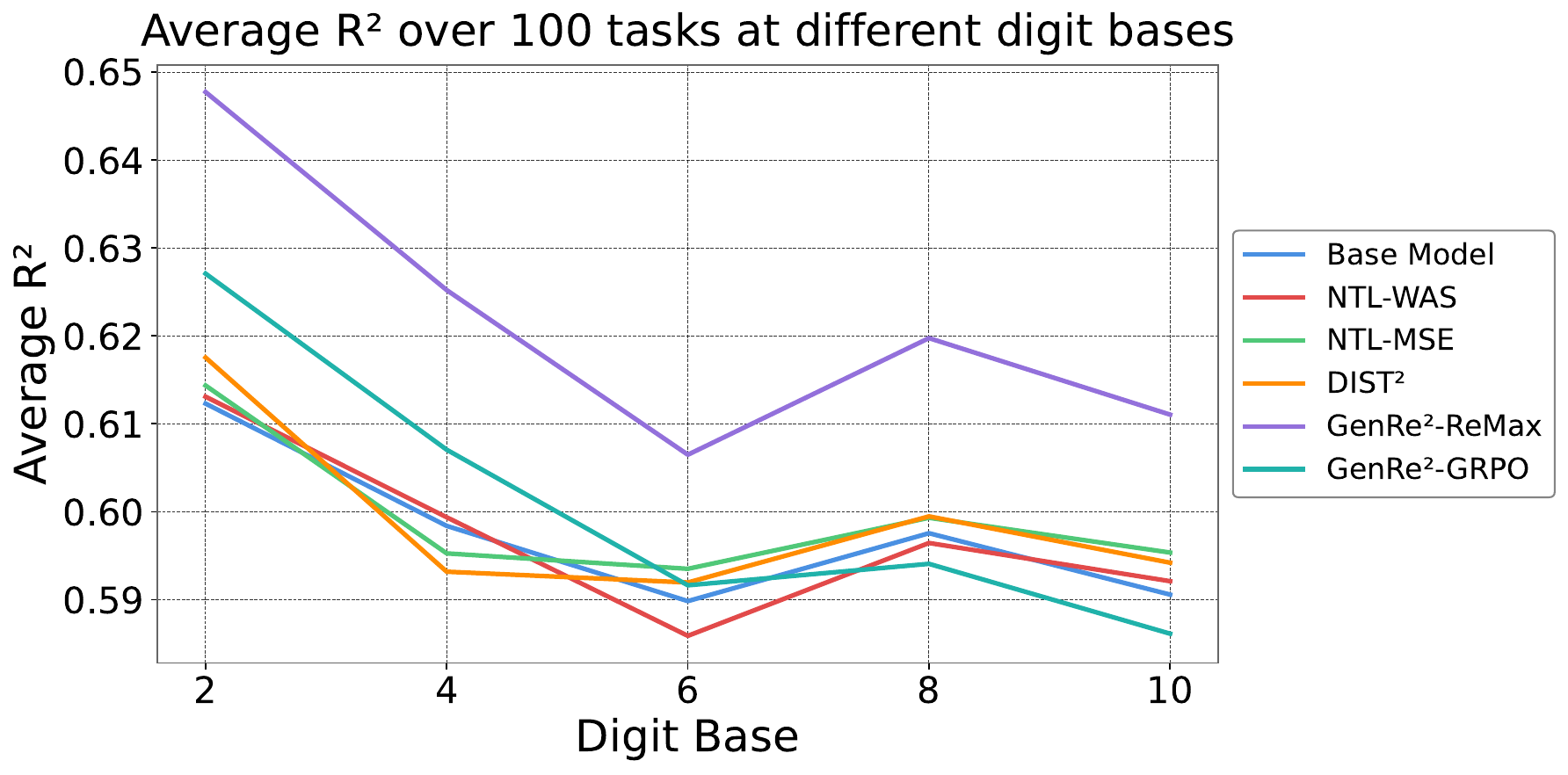}
    \vspace{-1.5em}
    \caption{Average R${}^2$ over 100 TALENT regression tasks of different methods under varying normalized tokenization digit bases.}
    \label{fig:digit-base}
    \vspace{-1.5em}
\end{figure}

\textbf{Different tokenization settings.}
To examine the robustness of our method, we vary the digit bases of the normalized tokenization from 2 to 10.
The results are illustrated in Figure~\ref{fig:digit-base}, where: (1) \ours-ReMax consistently achieves the highest R$^2$ scores across all digits bases, showing great robustness and improvements against token-level methods;
(2) In contrast, \ours-GRPO exhibits high sensitivity to this hyperparameter, and its performance degrades drastically as the digit base increases, even underperforming the base model at base 10.
To better understand this performance gap, we next conduct a case study to analyze these two RL methods. 
Additionally, we provide the ablation on different tokenizers in Appendix~\ref{Appdix:tokenizer}.

\begin{table}[t!]
\caption{Ablation study on the three key differences between \ours-GRPO and \ours-ReMax, averaging over 5 random seeds across 100 TALENT regression tasks. The best and runner-up results are \textbf{bolded} and \underline{underlined}, respectively. 
Rows shaded in \colorbox{gray!15}{gray} indicate experiments \textbf{with} Reward Standardization (Rew. Std.) enabled. IS Clip denotes Importance Sampling Clipping.
}
\vspace{0.3em}
\centering
\resizebox{\linewidth}{!}{
\begin{tabular}{l|ccc|c|c|c}
    \toprule
    & \multicolumn{3}{c|}{\textbf{Components}} & \multicolumn{3}{c}{\textbf{Metrics}} \\
    \cmidrule(lr){2-4} \cmidrule(lr){5-7}
    \multirow{-2}{*}{\textbf{Method}} & \textbf{IS Clip} & \textbf{Rew. Std.} & \textbf{Baseline} & \textbf{RMSE} $\downarrow$ & \textbf{R$^2$} $\uparrow$ & \textbf{Rank Corr.} $\uparrow$ \\ 
    \midrule
    
    \ours-ReMax & \xmark & \xmark & Greedy & \textbf{0.5464} & \textbf{0.6108} & \textbf{0.7860} \\ 
    \midrule
    
    \rowcolor{gray!15} \ours-GRPO & \cmark & \cmark & Mean & 0.5634 & 0.5872 & 0.7717 \\
    
    \rowcolor{gray!15} \quad $+$ Greedy Base. & \cmark & \cmark & Greedy & 0.5629 & 0.5876 & 0.7725 \\
    
    \rowcolor{gray!15} \quad $-$ IS Clip & \xmark & \cmark & Mean & 0.5637 & 0.5854 & 0.7717 \\
    
    \quad $-$ Rew. Std. & \cmark & \xmark & Mean & 0.5478 & 0.6089 & 0.7836 \\
    
    Greedy Variant & \cmark & \xmark & Greedy & \underline{0.5472} & \underline{0.6095} & \underline{0.7840} \\
    
    \bottomrule
\end{tabular}
}
\vspace{-1.5em}
\label{tab:ablation-remax&grpo}
\end{table}

\textbf{Ablation on \ours-GRPO components.}
To analyze the performance gap, we ablate the different components of GRPO and ReMax at digit base 10, where \ours-GRPO performs significantly worse than \ours-ReMax in Figure~\ref{fig:digit-base}. 
We note that GRPO differs from ReMax in three perspectives: 
(1) clipping important sampling ratio; 
(2) dividing reward by its standard deviation;
and (3) using the mean reward as the baseline value while ReMax uses greedy baseline.
As shown in Table~\ref{tab:ablation-remax&grpo}, reward standardization is the primary cause of \ours-GRPO's degradation.
We hypothesize this degradation results from biased gradient estimation from reward standardization~\citep{Dr-GRPO}, which hampers calibrated prediction~\citep{grpo-uncalibrated}.
However, as recent works also demonstrate the effectiveness of reward standardization in training stability, which can be viewed as an adaptive learning rate~\citep{grpo-normalization, llm-rl-survey-1}, we leave the discussion of this component as a future work.

\subsection{RLM for Code Metric Regression}
\label{sec:code}
In this subsection, we conduct experiments on Regression Language Model (RLM)~\citep{omnipred, regress_lm}, an important downstream application of decoding-based regression, to perform code metric regression, following~\citep{code_rlm}.
Specifically, we finetune the pretrained checkpoints provided by~\citep{code_rlm}~\footnote{\url{https://huggingface.co/akhauriyash/RLM-GemmaS-Code-v0}} on two datasets collected by~\citep{code_rlm}~\footnote{\url{https://huggingface.co/datasets/akhauriyash/Code-Regression}}~\footnote{We exclude the CodeNet dataset~\citep{codenets} currently due to its large scale.}: 
\begin{itemize}[leftmargin=1em, labelindent=0em]
    \item APPS Leetcode~\citep{apps}, which primarily involves predicting peak memory usage for high-level Python code, and the objectives include computational latency and memory usage;
    \item Triton Kernel Latency~\citep{kbss}, which focuses on estimating the execution latency of PyTorch programs for low-level Triton GPU kernels.
\end{itemize}

\subsubsection{Experimental Setup}
\label{sec:code-settings}
\textbf{Model architecture.}
The pretrained model provided by~\citep{code_rlm} is an encoder-decoder model, where the encoder is a pretrained T5Gemma~\citep{t5gemma} encoder and the decoder is a standard Transformer decoder trained from scratch with the IEEE tokenizer~\citep{IEEE} with digit base $B=10$, exponent length $E=3$, and mantissa length $M=5$. 
Since~\citet{code_rlm} trained the model with the encoder frozen, we also freeze the encoder and finetune the decoder with respective strategies.

\textbf{Compared methods.}
We consider decoding-based baselines for finetuning the given checkpoint, including finetuning by CE, NTL-WAS, NTL-MSE~\citep{ntl}, and DIST$^2$~\citep{DIST2}. 

\textbf{Training \& evaluation.} 
We randomly split the datasets (i.e., APPS Leetcode or Triton Kernel Latency) into training, validation, and test sets with proportions of 8:1:1, and finetune the model for 20 epochs using AdamW optimizer~\citep{adamw} with a learning rate of $1\times 10^{-6}$. 
We then evaluate the tuned model that achieves the best validation loss / reward, taking the median of $m=64$ generated samples for evaluation, following~\citep{code_rlm}. 

\begin{table*}[t!]
\vspace{-1.0em}
\caption{Results for code metric regression on APPS Leetcode and Triton Kernel Latency datasets comparing RMSE, R${}^2$ and Rank Correlation. Due to high training overhead, we train a single model and report results as the average over 5 random inference seeds. The best and runner-up results are \textbf{bolded} and \underline{underlined}, respectively. The row shaded in \colorbox{gray!15}{gray} indicates our proposed method.}
\centering
\resizebox{\linewidth}{!}{
\begin{tabular}{l|ccc|ccc}
    \toprule
        &\multicolumn{3}{c|}{APPS Leetcode} &\multicolumn{3}{c}{Triton Kernel Latency} \\ \cmidrule{2-7}
        \multirow{-2}{*}{Model}&  RMSE $\downarrow$ & R$^2$ $\uparrow$ & Rank Corr. $\uparrow$ &  RMSE  $\downarrow$& R$^2$ $\uparrow$ & Rank Corr. $\uparrow$ \\ \midrule
        Base Model &  \underline{0.493$_{\pm 0.000}$} & \underline{0.009$_{\pm 0.001}$} & \underline{0.935$_{\pm 0.000}$} & \underline{1.095$_{\pm 0.000}$} & \underline{-0.003$_{\pm 0.000}$} & 0.536$_{\pm 0.003}$ \\ \midrule
        +CE &  0.495$_{\pm \text{1.28}\!\times\!\text{10}^{\text{-6}} }$ & -0.002$_{\pm \text{5.19}\!\times\!\text{10}^{\text{-6}} }$ & 0.913$_{\pm 0.001}$ & 16.37$_{\pm 1.719}$ & -224.8$_{\pm 47.81}$ & \underline{0.555$_{\pm 0.001}$}  \\
        +NTL-WAS~\citep{ntl} &  0.495$_{\pm \text{2.20}\!\times\!\text{10}^{\text{-7}} }$ & -0.002$_{\pm \text{8.89}\!\times\!\text{10}^{\text{-7}} }$ & 0.904$_{\pm 0.001}$ & 23.99$_{\pm 1.625}$ & -481.6$_{\pm 64.20}$ & 0.539$_{\pm 0.010}$ \\
        +NTL-MSE~\citep{ntl} & 0.495$_{\pm \text{4.64}\!\times\!\text{10}^{\text{-7}} }$ & -0.002$_{\pm \text{1.88}\!\times\!\text{10}^{\text{-6}} }$ & 0.867$_{\pm 0.002}$ & 33.32$_{\pm 1.795}$ & -928.9$_{\pm 101.2}$ & 0.510$_{\pm 0.008}$ \\ 
        +DIST$^2$~\citep{DIST2} &  0.495$_{\pm \text{1.37}\!\times\!\text{10}^{\text{-6}} }$ & -0.002$_{\pm \text{5.56}\!\times\!\text{10}^{\text{-6}} }$ & 0.902$_{\pm 0.002}$ & 560.4$_{\pm 52.74}$ & ${\text{-2.64}\!\times\!\text{10}^{\text{5}}}_{\scriptstyle\pm \text{5.06}\times\text{10}^{\text{4}}}$ & 0.540$_{\pm 0.006}$ \\ \midrule
        \rowcolor{rowgray}\textbf{+\ours-ReMax (Ours)} &  \textbf{0.474}$_{\pm \text{5.41}\!\times\!\text{10}^{\text{-6}} }$ & \textbf{0.083}$_{\pm \text{2.10}\!\times\!\text{10}^{\text{-5}} }$ & \textbf{0.967}$_{\pm \text{7.34}\!\times\!\text{10}^{\text{-5}} }$ & \textbf{1.094}$_{\pm \text{8.44}\!\times\!\text{10}^{\text{-7}} }$ & \textbf{-0.001}$_{\pm \text{1.54}\!\times\!\text{10}^{\text{-6}} }$ & \textbf{0.598}$_{\pm 0.001}$ \\
\bottomrule
\end{tabular}
}
\label{tab:rlm-res}
\vspace{-1em}
\end{table*}

\textbf{RL details.}
We set the rollout budget $G=4$ in this experiment.
Before calculating the reward, we set the mapping $\psi$ as a quantile transformation towards a standard Gaussian distribution. 
The number of quantiles is adaptively set to $\operatorname{clip}(\lfloor N_{\textrm{train}}/{30}\rfloor, 10, 1000)$, where $N_{\textrm{train}}$ stands for the training set size.
We use the quantile transformation instead of z-score standardization to mitigate the impact of outliers on the reward. 
As shown in Figure~\ref{fig:quantile} in Appendix~\ref{appendix:code-normalization}, the objective distribution under z-score retains heavy tails and extreme values, while the quantile normalization suppresses outliers to yield a well-behaved Gaussian distribution.
Additionally, we clip the reward by a minimum negative MSE of $-50$ by: $R(\tau) = \max \left( -(\psi(\hat{y}) - \psi(y))^2, -50 \right)$.
We use \ours-ReMax~\citep{remax}, the best-performing method in Section~\ref{sec:tabular}, as our RL backbone.

\subsubsection{Results}
\label{sec:code-result}

Table~\ref{tab:rlm-res} summaries the results of different regression metrics on the two datasets.
\ours-ReMax achieves superior performance across all metrics, showing steady improvements against the base model. 
Notably, no individual token-level technique outperforms the base model after dataset-specific finetuning, which is also reported by~\citep{code_rlm}. 
We hypothesize this is a form of catastrophic forgetting, where specifically finetuning on the subset may negatively affect the general regression ability.
Instead, \ours can mitigate the forgetting compared to other token-level baselines, which aligns with the observation in LLM post-training that RL forgets less than SFT~\citep{the-path-not-taken, rl-forget-less-1, rl-forget-less-2}.

\subsection{Understanding the Effectiveness of RL for Decoding-based Regression}
\label{sec:analysis}
In this subsection, we conduct illustrative experiments on tabular regression tasks to understand the effectiveness of RL in decoding-based regression. 

In RLVR for LLM, \citet{limit-of-rlvr} showed that RL-tuned models do not exceed the potential of the base model.
They found that under the standard implementation, RL often reduces the model's reasoning capacity by observing that the base model often outperforms the RL-tuned model on pass@$k$ at large $k$, a widely adopted metric for RLVR~\citep{pass-at-k, pass-at-k-1} measuring the probability of obtaining at least one correct solution in $k$ independent samples. But RL significantly improves sampling efficiency by boosting pass@1~\citep{limit-of-rlvr, rl-squeeze}, thus showing great capability in real-world application.

However, standard regression metrics derived from aggregation (e.g., mean or median) just reflect the expected utility but mask the capability boundary (i.e., the potential to generate a precise solution), which is different from pass@$k$.
Therefore, to disentangle these factors and probe the theoretical limit of the model's capacity, we analyze the best@$k$ metric under an oracle selection setting.
Specifically, for a given feature $\phi(\x_{i})$ in the test set, the model generates $k$ predictions $\{ \hat{y}_{i}^1, \dots, \hat{y}_{i}^k \}$ and selects the closest one to the ground truth $y_i$: 
\begin{equation*}
    \hat{y}_{i}^{\text{best}} = \mathop{\arg\min}\nolimits_{\hat{y} \in \{\hat{y}_{i}^1, \dots, \hat{y}_{i}^k\}} |y_i - \hat{y}|.
\end{equation*}
Then, the best@$k$ metrics can be calculated using the collection of all $\hat{y}_{i}^{\text{best}}$ values.

\begin{figure}[!t]
    \centering
    \includegraphics[width=\linewidth]{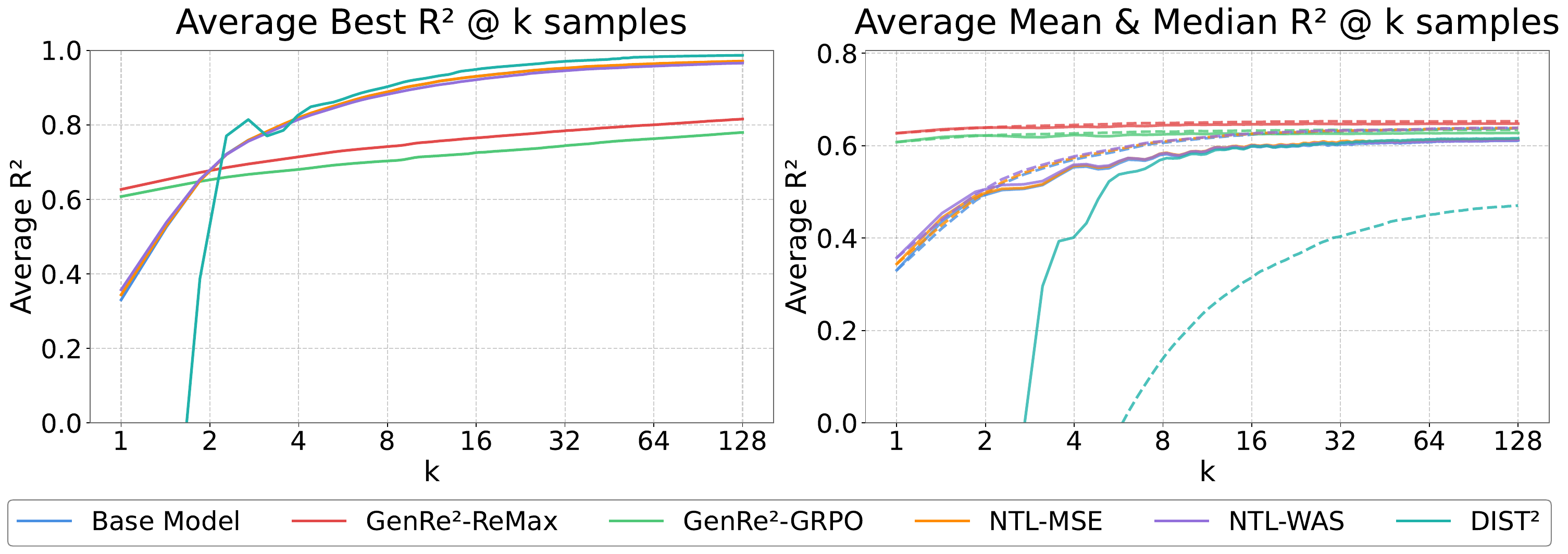} %
    \vspace{-1.5em}
    \caption{Metric dynamics across 100 TALENT regression tasks. The left sub-figure displays the average best R${}^2$@$k$, while the right one shows the average mean (dashed) and median (solid) R${}^2$@$k$.}
    \label{fig:best-of-k}
    \vspace{-1em}
\end{figure}

\begin{figure}[!t]
    \centering
    \includegraphics[width=\linewidth]{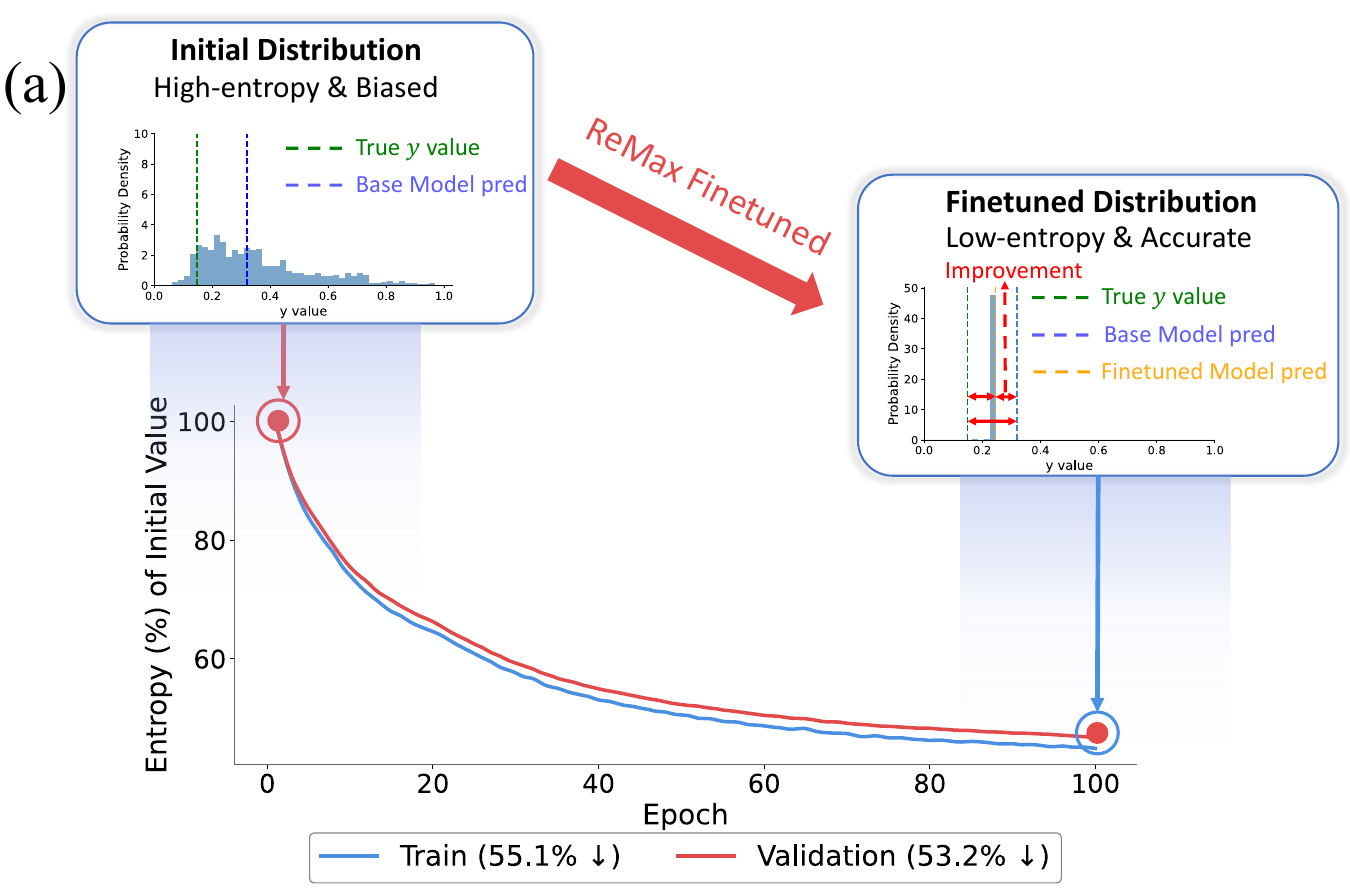} \\
     \includegraphics[width=\linewidth]{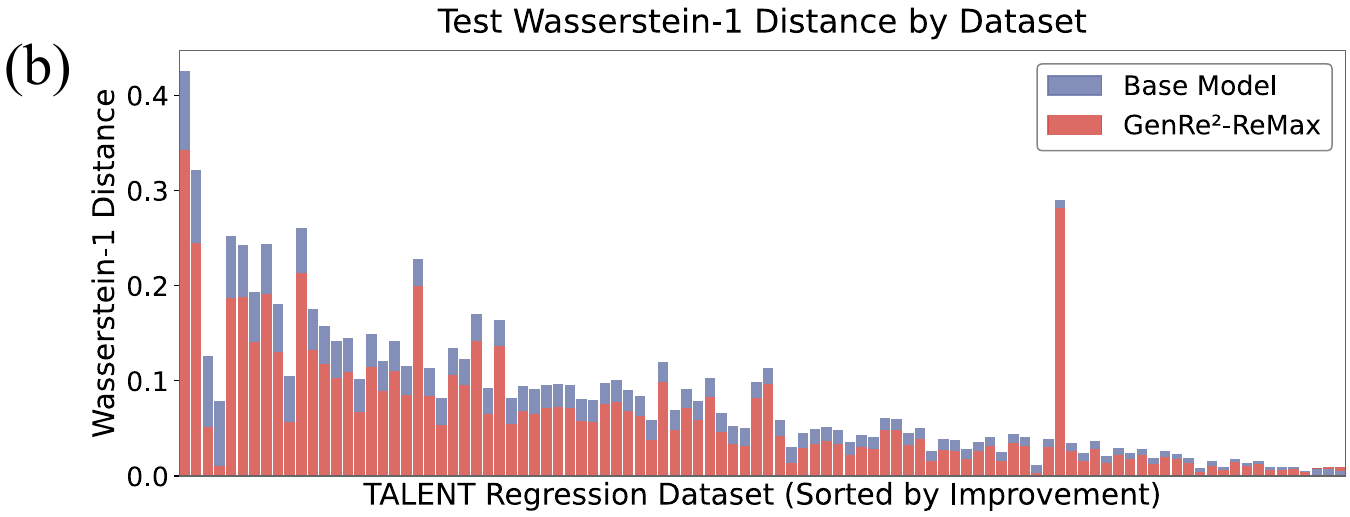}
    
    \vspace{-0.5em}
    \caption{Impact of \ours finetuning on output distribution.
    \textbf{(a)} \ours significantly reduces entropy during training, transforming the initial high-entropy distribution into a sharper, low-entropy distribution that is more accurate  (visualized on the \texttt{Kaggle\_\allowdisplaybreaks bike\_\allowdisplaybreaks sharing\_\allowdisplaybreaks demand\_\allowdisplaybreaks challange}~\citep{kaggle} task).
    \textbf{(b)} Visualization of the test Wasserstein-1 distance (lower is better) across 100 regression datasets of TALENT benchmark, where \ours-ReMax (red) consistently achieves lower distances compared to the base model (blue), demonstrating better approximation towards the ground truth target.
    }
    \vspace{-1.5em}
    \label{fig:output-distribution}
\end{figure}

We plot the best R$^2$@$k$ with varying $k$ in the left sub-figure of Figure~\ref{fig:best-of-k}.
Consistent with the observations in RLVR~\citep{limit-of-rlvr, NSR}, we find that the \ours-tuned models significantly improve best R$^2$@1 but surpassed by the base model as $k$ increases.
While all token-level finetuning methods maintain performance at large $k$, \ours implicitly suppresses exploration of the output space, thereby lowering the capability boundary.
However, this lower variance allows \ours to generate better solutions in a single trial (i.e., better best@1), thus achieving better mean and median regression performance shown in the right sub-figure of Figure~\ref{fig:best-of-k}. 

To better understand the effectiveness of \ours, we visualize the evolution of the model's output distribution in Figure~\ref{fig:output-distribution}.
In (a), the average entropy drops by over 50\% during training, transforming initial high-entropy, biased distributions into sharper, more accurate predictions.
Additionally, Figure~\ref{fig:output-distribution} (b) shows that \ours-ReMax consistently achieves lower test Wasserstein-1 distances towards the target value than the base model on most tasks, demonstrating a better approximation of the ground truth target.

\section{Discussion}
\textbf{Conclusion.} 
In this paper, we emphasize the significance of decoding-based regression. 
We challenge the practice of training the model via token-level loss, and propose \ours to address the limitations of prior methods by utilizing Reinforcement Learning. 
Experimental results on tabular regression and code metric regression show the superiority and generalization of \ours, demonstrating the effectiveness of sequence-level reward signals overall token-level supervisions.

\textbf{Future works.}
Based on our experimental results and analyses, there are many worthwhile directions for future exploration. 
Here we highlight some promising avenues for future research:
\begin{enumerate}[leftmargin=1em, labelindent=0em]
    \item[1.] \textbf{Extending to generative reward models and verifiers.} 
    Decoding-based regression is relevant to Generative Reward Models (GRMs), which also score the inputs in an end-to-end manner.
    While current works have introduced RL to GRMs~\citep{GenRM-2, scaling-verifier, Heimdall}, they primarily rely on sparse ranking signals based on the final solution without analyzing the intermediate procedure.
    Notably, the recently proposed DeepSeek-Math-V2~\citep{deepseek-math-v2} introduces regression-like rewards for verifier RL training. 
    It remains to be studied whether \ours can be effectively extended to enhance the performance of RL-trained regression-based verifiers.
 
\item[2.]\textbf{Robust uncertainty calibration.}
Our experiments in Section~\ref{sec:analysis}, together with prior works in RLVR~\citep{NSR, limit-of-rlvr, rl-squeeze} indicate that while RL is effective, it tends to over-sharpen the output distribution, which leads to uncalibrated prediction.
However, this harms the uncertainty estimation capability delivered from pretraining~\citep{decoding_regression, grpo-uncalibrated}, which is important for response verification~\citep{p1} and real-world usage~\citep{TNP, uncertainty, embed-then-regress}. 
Thus, an urgent need still exists for developing more robust and calibrated generative decoding-based regressors under the dynamics of RL-based post-training.

\item[3.]\textbf{Understanding the mechanism of RL update.}
Recent works have identified the sparse weight update dynamics of RLVR~\citep{who-reason-in-llm, rl-small-subnet, the-path-not-taken}, which motivates for geometry-aware, parameter-efficient RL algorithm design.
Although RL shows consistent improvements in decoding-based regression, the underlying mechanisms remain underexplored.
It is worth studying further how RL changes the parameter and its association with regression metrics.

\item[4.]\textbf{Better RL algorithms.}
Although RLVR algorithms (i.e., ReMax and GRPO used in this paper) show good capabilities, our analysis of the best$@k$ metric in Section~\ref{sec:analysis} suggests that current algorithms have not fully explored the search space for decoding-based regression.
Thus, techniques like entropy regularization~\citep{entropy-mechanism}, improving sampling efficiency~\citep{RLsharp}, and negative samples reinforcement~\citep{NSR} can be further explored.

\item[5.] \textbf{Combination with modern tabular regression structures.}
In this paper, we mainly use MLP as the encoder for tabular regression. 
The generalization of decoding-based regression upon other prevalent tabular model structures~\citep{t2g-former,excelformer,FT-Transformer,modernnca} and tabular foundation models~\citep{limix, tabpfn, tabpfn-2.5, mitra} is worth further studied.

\end{enumerate}

\bibliography{example_paper}
\bibliographystyle{icml2025}

\newpage
\appendix
\onecolumn

\section{Additional Backgrounds}

\subsection{Regression}
\label{sec: reg}
Given a training dataset $\dtrain = \{ (\x_i, y_i) \}_{i=1}^N$ sampled from an unknown ground-truth function $f:\mathcal{X} \rightarrow \mathbb{R}$, regression aims to learn a model from $\dtrain$ that accurately predicts the output for unseen inputs. 
The quality of the learned model is evaluated on a hold-out test set, $\dtest$, by measuring its predictive ability with regression-based metrics, e.g., Root Mean Squared Error (RMSE).

Traditional regression methods involve statistical techniques like Gaussian Processes~\citep{gpml} and tree-based methods~\citep{xgboost, deep-forest}. 
Recent works have focused on deep learning (DL)-based methods~\citep{FT-Transformer, dl-tabular-survey}, which train deep neural networks to leverage the power of representation learning for regression, demonstrating great superiority and scalability~\citep{FT-Transformer, talent-2}.
Specifically, DL-based methods train neural networks to map the input $\x$ to a high-dimensional representation $\phi(\x)$, and subsequently models the probability distribution of the target value $p_{\bds{\theta}}(y\mid\phi(\x))$ via a regression head, parameterized by $\bds{\theta}$.

There are several design philosophies for the regression head, including \textit{pointwise head}, \textit{parametric distribution head}, and \textit{Riemann head}. 
The \textit{pointwise head} maps $\phi(\x)$ to a scalar prediction, which is the most commonly used regression head. 
However, the pointwise head fails to capture both the uncertainty~\citep{uncertainty-1} and the complex multimodality of the target distribution~\citep{bishop1994mixture}.
To address this, the \textit{parametric distribution head} instead models output as a predefined distribution (e.g., Gaussian) and predicts its parameters (e.g., the mean value and the standard variance)~\citep{CNP, NP, TNP}. 
The \textit{Riemann head}, also called \textit{histogram head}, instead converts the regression problem into classification by discretizing the continuous output $y$ into finite bins~\citep{histogram-head}.
The learned model predicts the probability of each bin, from which the output value is derived using a weighted sum. 
Though sensible to hyperparameters~\citep{hist-hpm-sensible}, the \textit{Riemann head} has been shown to improve the model's robustness and performance~\citep{histogram-head, histogram-head-2}, with successful application to reinforcement learning (RL)~\citep{hist-rl-1, hist-rl-2} and tabular foundation models~\citep{tabpfn, tabpfn-2.5, limix}.

\subsection{RL for LLM}
\label{appendix:RL4LLM}

Reinforcement Learning (RL) has become a pivotal post-training technique for LLMs~\citep{llm-rl-survey-1}, popularized by RLHF for alignment~\citep{RLHF-0, RLHF-1, GPT4}, and extended to domains with verifiable rewards like mathematical~\citep{deepseek-r1, llm-math-survey} and scientific reasoning~\citep{intern-s1, p1}. 
RL approaches are primarily categorized into offline preference optimization~\citep{DPO, SimPO, KTO} and online policy gradient methods~\citep{PPO, reinforce}. 
While early online methods relied on actor-critic algorithms like PPO~\citep{PPO}, recent works leverage the deterministic transitions of LLMs to adopt lightweight REINFORCE-based methods without a value model~\citep{reinforce}. 
Notably, ReMax~\citep{remax} and GRPO~\citep{GRPO} reduce variance using greedy and multi-sample mean baselines, respectively, with the latter showing power in DeepSeek-R1~\citep{deepseek-r1}. 
Further variants enhance scalability and stability through reducing variance~\citep{reinforce++} and estimation biases~\citep{RLOO, Dr-GRPO}, and employing regularization techniques such as entropy~\citep{entropy-mechanism} and reward shaping~\citep{pass-at-k-training, SimKO}. 
Compared to Supervised Finetuning (SFT), RL demonstrates superior generalization~\citep{rl-generalize, rl-squeeze, the-path-not-taken} and mitigated forgetting~\citep{rl-forget-less-1, rl-forget-less-2}. 

\section{Description of Tokenizations}
\label{sec:tokenization}
Here we provide detailed descriptions of the two common tokenization strategies introduced in the main text:
\begin{enumerate}
    \item [$\bullet$] \textit{Normalized Tokenization}: 
    The normalization tokenization first scales a target value $y$ to a fixed interval (e.g. $[0, 1]$). 
    Then, this method represents the scaled value as a base-$B$ expansion. 
    For instance, when choosing $B=2$ and a mantissa length of $M=3$, the scaled number $0.6$ is tokenized as \texttt{<1><1><0>}. 
    For prediction, we need rescale the detokenized number to its original space.
    However, normalized tokenization relies on the access to the global minimum and maximum.
    One could set $y_{\min}$ and $y_{\max}$ in accordance with the training dataset, but this method is highly sensitive to outliers~\citep{power-transform}. 
    Besides, it is unsuitable for multi-task regression, where different tasks may have different objectives, as globally linear scaling to $[0,1]$ can cause precision loss~\citep{omnipred}.
    \item[$\bullet$] \textit{Scientific Notation Tokenization}:
    Unlike normalized approaches, scientific notation tokenization methods do not normalize the target value. 
    Instead, they represent numbers using sign, mantissa, and exponent components. We describe two specific implementations below:
    \begin{itemize}
        \item[$\bullet$] \textit{P10 Tokenization}~\citep{P10}: 
        P10 Tokenization is an unnormalized tokenization method that represents numbers in a format similar to scientific notation. 
        It breaks down a scalar into three components: A sign token, a mantissa part with $M$ tokens, and an exponent token. 
        For example, with a mantissa length of $M=3$, the number $1.23$ is tokenized as \texttt{<+><1><2><3><E-2>}.
        \item[$\bullet$] \textit{IEEE Tokenization}~\citep{IEEE}: IEEE Tokenization is another unnormalized tokenization scheme that directly represents a target value $y$ by generalizing the IEEE-754 floating-point standard into a base-$B$ format. 
        It tokenizes a number into a sequence representing its sign, exponent and mantissa. 
        For instance, with base $B=10$, an exponent length of $E=3$, and a mantissa length of $M=4$, the number $10^{-12} \times 1.234$ is tokenized as \texttt{<+><-><0><1><2><1><2><3><4>}.
    \end{itemize}
\end{enumerate}

\section{Policy Gradient Methods}
\label{sec:pg_methods}
To optimize the objective in Eq.~\eqref{eq:objective}, we employ the policy gradient method~\citep{pg-method} to optimize $\pi_{\bds\theta}$ by ascending the gradient of the expected return:
\begin{equation}%
\begin{aligned}
    \nabla_{\bds\theta}\mathcal{J}(\bds\theta)=\mathbb{E}_{(\x, y) \sim \dtrain }\mathbb{E}_{\tau\sim\pi_{\bds\theta}} \left[ \sum_{k=0}^{K-1} \nabla_{\bds\theta} \log \pi_{\bds\theta}(a_k\mid s_k)A^{\pi_{\bds\theta}}(s_k, a_k) \right],
\end{aligned}
\end{equation}
where $A^{\pi_{\bds\theta}} (s_k, a_k)$ is the advantage function estimating the relative value of action $a_k$ in state $s_k$. 
As the expectation $\mathbb{E}_{\tau\sim\pi_{\bds\theta}}$ is intractable, one practical solution is to approximate it via Monte Carlo sampling.

Given the deterministic state transitions in the MDP, simple policy gradient methods like REINFORCE~\citep{reinforce} are efficient.
To reduce variance, REINFORCE subtracts a baseline in the advantage function:
\begin{align*}
    A^{\pi_{\bds\theta}}(s_k,a_k)=R(\tau) - b(\phi(\x)),
\end{align*}
where $\tau=(s_0, a_0,\cdots,s_K)$ denotes a trajectory sampled from $\pi_{\bds\theta}$, $R(\tau)=\sum_{k=0}^{K-1} r(s_k,a_k)$ is the expected return of the trajectory $\tau$, and $b(\phi(\x))$ is the baseline value related to the input $\phi(\x)$, which is to be designed. 
Crucially, this subtraction maintains an unbiased estimator~\citep{reinforce}, forming the foundation of online policy gradient methods~\citep{rl-book}.

In this paper, we employ two prevalent algorithms, ReMax~\citep{remax} and GRPO~\citep{GRPO}, which can be viewed as REINFORCE variants with distinct advantage formulations. 
ReMax reduces variance efficiently by setting the baseline as the reward of a greedy decoding sequence:
\begin{align*}
    A_{\textrm{ReMax}}^{\pi_{\bds\theta}}(\tau) = R(\tau)-r(\phi(\x),\hat{a}_{0:K-1}), \text{ where } \hat{a}_k\in \mathop{\arg\max} \pi_{\bds\theta}(\cdot\mid\phi(\x), \hat{a}_{0:k-1}).
\end{align*}
GRPO, on the other hand, computes the advantage by normalizing rewards relative to a group of $G$ sampled trajectories $\{\tau^i\}_{i=1}^G$:
\begin{align*} 
     A^{\pi_{\bds\theta}}_{\textrm{GRPO}}(\tau^i) = \frac{R(\tau^i) - \operatorname{mean}_j\left\{R(\tau^j)\right\}}{ \operatorname{std}_j\left\{R(\tau^j)\right\} + \epsilon},
\end{align*} 
where $\epsilon$ is a small positive constant. Additionally, GRPO further stabilizes training by incorporating importance sampling and clipping mechanisms into its final objective, which is defined as:
\begin{equation*}%
\begin{aligned}
    \mathcal{J}(\bds\theta) =& \mathbb{E}_{{\tau \sim \pi_{\bds\theta}}} \Bigg[  \frac{1}{G} \sum_{i=1}^G \min\left\{\mathrm{IS}(\bds{\theta})A_{\textrm{GRPO}}^{\pi_{\bds\theta}}(\tau^i),  \operatorname{clip}(\mathrm{IS}(\bds\theta), 1-\varepsilon, 1+\varepsilon)A_{\textrm{GRPO}}^{\pi_{\bds\theta}}(\tau^i) \right\}\Bigg],
\end{aligned}
\end{equation*}
where $\mathrm{IS}(\bds\theta) = \frac{\pi_{\bds\theta}(\tau^i\mid\phi(\x))}{\pi_{\bds\theta_{\text{old}}}(\tau^i\mid\phi(\x))}$ denotes the importance sampling ratio between the current and reference policies, and $\varepsilon$ is a hyperparameter controlling the clipping range.

\section{Experimental Settings}
\label{sec:appendix_settings}

\subsection{Baseline Details}
\label{sec:baseline-details}
In this section, we provide detailed implementations of the baselines compared in our experiments. 
Consistent with the main text, we categorize these methods into two groups: (1) \textbf{Baselines with different regression heads}, specifically the Pointwise head and the Riemann head; 
and (2) \textbf{Decoding-based regression methods}, which incorporate token-level loss improvements including NTL variants (NTL-MSE, NTL-WAS) and DIST$^2$.
\subsubsection{Pointwise Head}
The Pointwise head represents the standard regression approach. 
It projects the latent representation $\phi(\x)$ directly to a scalar prediction $\hat{y} \in \mathbb{R}$ via a linear layer. 
The model is optimized by minimizing the Mean Squared Error (MSE) loss between the predicted value and the ground truth:\begin{align*}
    \mathcal{L}_{\mathrm{MSE}}=\frac{1}{N} \sum_{i=1}^{N}\left(y_{i}-\hat{y}_{i}\right)^{2},
\end{align*}
where $N$ denotes the number of samples in the batch, $y_{i}$ and $\hat{y}_{i}$ denote the true value and the predicted value, respectively.

\subsubsection{Riemann Head}
Following~\citep{pfn,pfns4bo}, we implement the Riemann head by combining the infinite support architecture with the histogram loss objective~\citep{histogram-head}. 
This approach models the regression target as a probability distribution rather than a single scalar, allowing for better handling of uncertainty and outliers.

\textbf{Infinite Support Architecture.} 
We partition the target space of $y$ into a central finite range and two infinite tails to handle potential outliers. 
The central range $[y_{\text{min}}, y_{\text{max}}]$ is divided into $K$ uniform bins, each with width $w = (y_{\text{max}} - y_{\text{min}}) / K$. 
Additionally, we define a left tail region for $y < y_{\text{min}}$ and a right tail region for $y \ge y_{\text{max}}$. 
The neural network outputs a probability vector $\boldsymbol{o} = [o_0, \dots, o_{K+1}]$, representing the probability mass assigned to each component. 
The full predictive Probability Density Function (PDF), denoted as $q_{\bds\theta}(y\mid \x)$, is defined piecewise:
\begin{equation*}
q_{\bds\theta}(y \mid \x)=\left\{\begin{array}{ll}\underbrace{\frac{o_{0}}{\sigma_{\text {tail }}} \phi_{H N}\left(\frac{y_{\min }-y}{\sigma_{\text {tail }}}\right)}_{\text {Left Tail (Half-Normal) }} & \text { if } y<y_{\min } \\ \underbrace{\frac{o_{k}}{w}}_{\text {Central Bins (Uniform) }} & \text { if } y \in\left[y_{\min }+(k-1) w, y_{\min }+k w\right), k \in \{1,2,\ldots,K\} \\ \underbrace{\frac{o_{K+1}}{\sigma_{\text {tail }}} \phi_{H N}\left(\frac{y-y_{\max }}{\sigma_{\text {tail }}}\right)}_{\text {Right Tail (Half-Normal) }} & \text { if } y \geq y_{\max }\end{array}\right.
\end{equation*}
where $\phi_{HN}(\cdot)$ is the PDF of the standard Half-Normal distribution, and $\sigma_{\text{tail}}$ is a fixed scale parameter (set to 0.5) controlling the decay rate in the tail regions.

\textbf{Histogram Loss.}
To train the model, we construct a smoothed target distribution. 
Given a ground truth scalar $y_{gt}$, we model the target distribution $p(y)$ as a Gaussian centered at $y_{gt}$ with standard deviation $\sigma = 0.75w$, truncated to the central range: $p(y) \propto \mathcal{N}(y; y_{gt}, \sigma^2) \cdot \mathbb{I}{[y_{\text{min}}, y_{\text{max}}]}$.
We then discretize this continuous target by integrating $p(y)$ over each bin's interval to obtain the target probability mass $p_k = \int_{l_k}^{r_k} p(y) dy$, where $[l_k, r_k]$ denotes the interval of the $k$-th central bin. 
The model is optimized by minimizing the CE loss between the target mass vector $\mathbf{p}$ and the predicted mass vector $\boldsymbol{o}$:\begin{equation*}
    \mathcal{L} = - \sum_{k=0}^{K+1} p_k \log(o_k).
\end{equation*}

\textbf{Inference.}
During inference, we obtain the final scalar prediction $\hat{y}$ by calculating the expected value of the predicted distribution $q_{\bds\theta}(y|x)$. 
This is computed as the weighted sum of the centroids of all components:
\begin{equation*}
\hat{y} = \mathbb{E}_{y \sim q_{\bds\theta}}[y] = \sum_{k=0}^{K+1} o_k \cdot c_k,
\end{equation*}
where $c_k$ is the centroid of the $k$-th component. For central bins, $c_k$ is the midpoint of the interval; for the tail regions, $c_k$ is the expectation of the shifted Half-Normal distribution.

\subsubsection{Number Token Loss (NTL)}
Number Token Loss (NTL) ~\citep{ntl} is an auxiliary regression objective designed to improve numerical predictability of autoregressive language models. 
Unlike standard CE, which treats numbers as independent nominal tokens, NTL penalizes the numerical distance between the predicted distribution and the ground truth of each numeric token. 
We implement the two primary variants proposed by the authors:

\textbf{NTL-MSE.}
This variant treats the model’s output as a continuous expectation. It minimizes the MSE between the numerical value of the ground truth token and the expected numerical value derived from the predicted probability distribution. Let $V(j)$ denote the numerical value of token $j$, $\omega_t$ be ground truth numeric token at step $t$, and $\mathcal{N}$ be the set of indices for number tokens:
\begin{align*}
\mathcal{L}_{\mathrm{NTL}-\mathrm{MSE}}=\frac{1}{K} \sum_{t=1}^{K}\left(V(\omega_{t})-\sum_{j \in \mathcal{N}} p_{t}^{j} \cdot V(j)\right)^{2},
\end{align*}
where $p_{t}^{j}$ denotes the predicted probability assigned to token $j$ at step $t$, , and $K$ represents the total number of numeric tokens in the sequence.

\textbf{NTL-WAS.}
To address potential optimization issues in MSE (e.g., non-unique minima), NTL-WAS minimizes the Wasserstein-1 distance. 
For a one-hot ground truth distribution, this simplifies to the expected absolute difference:\begin{align*}
    \mathcal{L}_{\mathrm{NTL}-\mathrm{WAS}}=\frac{1}{K} \sum_{t=1}^{K} \sum_{j \in \mathcal{N}} p_{t}^{j} \cdot\left|V(\omega_t)-V(j)\right|.
\end{align*}

\textbf{Implementation Details.}
The model is optimized using a joint objective: $\mathcal{L} = \mathcal{L}_{\text{CE}} + \lambda \cdot \mathcal{L}_{\text{NTL}}$, where we set the hyperparameter $\lambda$ to 0.3. 
The auxiliary loss is computed exclusively on numerical tokens, while non-numerical tokens are masked out.

\subsubsection{DIST${}^2$ Loss}
DIST${}^2$ Loss~\citep{DIST2} introduces a distance-aware framework that integrates metric relationships directly into the target distribution of discrete autoregressive models. 
DIST${}^2$ constructs a soft, categorical target distribution $p_d$ based on the inherent distance metric $d$ between the ground truth token $\omega$ and vocabulary tokens $j$.

The target distribution is modeled as a discretized exponential family distribution, where tokens closer to the ground truth in the metric space are assigned higher probabilities. This is controlled by a temperature parameter $T$:\begin{align*}
    p_{d}(j \mid \omega)=\frac{\exp (-d(j, \omega) / T)}{\sum_{j^{\prime} \in \mathcal{N}} \exp \left(-d\left(j^{\prime}, \omega\right) / T\right)},
\end{align*}
where we set $T=1.0$ and the distance metric $d$ as Euclidean distance. The objective minimizes the KL divergence between this distance-aware target distribution and the model's predicted distribution $p_{\bds\theta}$:\begin{align*}
    \mathcal{L}_{\mathrm{DIST} ^2}=\sum_{t=1}^{K} \sum_{j \in \mathcal{N}} p_{d}\left(j \mid \omega_{t}\right) \log \frac{p_{d}\left(j \mid \omega_{t}\right)}{p_{\bds\theta}\left(j \mid \omega_{<t}\right)},
\end{align*}
where $K$ denotes the sequence length, $\omega_{t}$ represents the ground truth token at step $t$, and $\mathcal{N}$ is the set of indices corresponding to number tokens in the vocabulary.

In this work, we adopt a joint training strategy that combines the standard CE loss ($\mathcal{L}_{\text{CE}}$) with the DIST${}^2$. The final optimization objective is formulated as: $
\mathcal{L}=\mathcal{L}_{\mathrm{CE}}+\lambda \cdot \mathcal{L}_{\mathrm{DIST} ^2}$.
Following the default configuration of the original paper, we set the weighting hyperparameter $\lambda$ to 0.1. The auxiliary loss is computed exclusively on numerical tokens, while non-numerical tokens are masked out.

\subsection{Model Architecture}
\label{sec:architecture}

Consistent with the decoding-based regression paradigm, all models employed in this work utilize an encoder-decoder framework. The specific architectural choices are tailored to the input modalities of the respective tasks.

\subsubsection{Tabular Regression}
\paragraph{Decoding-based Regressor.} 
Following \citep{decoding_regression}, we utilize a hybrid architecture composed of an MLP encoder and a Transformer decoder to handle numerical feature inputs.
\begin{itemize}
    \item \textbf{Encoder:} The encoder is implemented as a Multi-Layer Perceptron (MLP) to project continuous input features into the latent space. It consists of three hidden layers, each with a dimensionality of $1024$ and Rectified Linear Unit (ReLU) activation functions. The input layer dynamically adjusts to the dimensionality of the feature vector $\mathbf{x}$, while the final linear layer projects the representation to a model dimension of $d = 256$.
    
    \item \textbf{Decoder:} The decoder follows a standard Transformer architecture~\citep{transformer} to autoregressively generate the target token sequence. The model dimension is set to $d = 256$ to align with the encoder's output. The network comprises a stack of $3$ decoder layers with multi-head attention (the number of head is set to 4), balancing computational efficiency with modeling capacity.
    \item \textbf{Tokenizer Configuration:} For the decoding-based heads, we apply specific tokenization settings. For the P10~\citep{P10} and IEEE~\citep{IEEE} tokenizers, the default configuration preserves $4$ decimal places for precision, with an exponent length (order) of $10$.
\end{itemize}

\paragraph{Baseline Regressors.} 
We configure the baseline regression heads as follows:
\begin{itemize}
    \item \textbf{Pointwise Head (MLP):} To ensure a fair comparison, we scale the Pointwise regression baseline to have a parameter count comparable to the decoding-based model. Specifically, we implement it as a large MLP consisting of $3$ hidden layers, each with a dimensionality of $2048$ and ReLU activations. Such baseline setting ensures that the pointwise baselines have more parameter than the decoding-based regressors.
    
    \item \textbf{Riemann Head:} For the Riemann head baseline, we discretize the target space into $K=256$ bins. The support range for the bins is set to $[-3, 3]$ (applied to the normalized targets), with infinite tails handling values outside this range.
    The MLP encoder setting follows the same setting of pointwise head.
\end{itemize}
\subsubsection{Code Metric Regression}
We utilize the pretrained model provided by~\citep{code_rlm}~\footnote{\url{https://huggingface.co/akhauriyash/RLM-GemmaS-Code-v0}}, which is an encoder-decoder model. 
The encoder of the pretrained model is a pretrained T5Gemma~\citep{t5gemma} encoder and the decoder is a standard Transformer decoder trained from scratch with the IEEE tokenizer~\citep{IEEE}, configured with digit base $B=10$, exponent length $E=3$, and mantissa length $M=5$. 
Since~\citet{code_rlm} trained the model with the encoder frozen, we also freeze the encoder and finetune the decoder with respective strategies.

\subsection{Data Processing}
\label{appendix:norm_head}

We apply specific data processing strategies according to the input modality and the regression head employed. All statistics used for normalization are computed exclusively from the training set to prevent data leakage.

\paragraph{Input Processing.} 
The preprocessing of input features $\mathbf{x}$ depends on the task type:
\begin{itemize}
    \item \textbf{Tabular Regression:} Since the inputs are numerical vectors, we apply standard z-score normalization to enhance numerical stability:
    \begin{equation*}
        \mathbf{x} \leftarrow \frac{\mathbf{x} - \boldsymbol{\mu}_x}{\boldsymbol{\sigma}_x},
    \end{equation*}
    where $\boldsymbol{\mu}_x$ and $\boldsymbol{\sigma}_x$ denote the coordinate-wise mean and standard deviation of the training inputs, respectively.
    \item \textbf{Code Metric Regression:} The inputs for these tasks are raw textual code. Consequently, we feed the text directly into the encoder without applying any additional numerical normalization.
\end{itemize}

\paragraph{Target Processing.} 
The processing of target values $y$ is determined by the specific regression head and tokenization scheme:
\begin{itemize}
    \item \textbf{Non-Decoder Heads (Pointwise \& Riemann):} For these baselines, we standardize the targets using z-score normalization:
    \begin{equation*}
        y \leftarrow \frac{y - \mu_y}{\sigma_y},
    \end{equation*}
    where $\mu_y$ and $\sigma_y$ represent the mean and standard deviation of the training targets, respectively.
    
    \item \textbf{Decoder Heads:} For decoding-based regression, the strategy varies by tokenization scheme:
    \begin{itemize}
        \item \textit{P10 and IEEE Tokenization:} These schemes are designed to represent the raw numbers directly via scientific notation. Therefore, we do not apply any normalization and train the model to regress the raw target values.
        \item \textit{Normalized Tokenization:} This scheme requires targets to be bounded within a fixed interval. We adopt a two-stage scaling strategy: targets are first standardized via z-score, followed by Min-Max scaling. The transformation is defined as:
        \begin{equation*}
            y' = \frac{y - \mu_y}{\sigma_y}, \quad
            y \leftarrow \frac{y' - \min(y')}{\max(y') - \min(y')}.
        \end{equation*}
    \end{itemize}
\end{itemize}

\subsection{Implementation Details}
\label{sec:exp_detail}

\paragraph{Training Hyperparameters.} 
Optimization is performed using the AdamW optimizer. The specific configurations for each domain are as follows:
\begin{itemize}
    \item \textbf{Tabular Regression:} For decoding-based methods, we use a batch size of $128$ and an initial learning rate of $1 \times 10^{-5}$. The learning rate follows a cosine annealing schedule with $100$ warmup steps and a minimum decay ratio of $0.1$. The Base Model pretraining (CE) is conducted for $200$ epochs, while the proposed Policy Gradient optimization runs for $100$ epochs. For the baseline regression heads (MLP), we utilize the default training framework and hyperparameters provided by the TALENT benchmark.
    \item \textbf{Code Metric Regression:} We employ a batch size of $16$ with a lower initial learning rate of $1 \times 10^{-6}$ to preserve the pre-trained knowledge of the backbone. The model is fine-tuned for a total of $20$ epochs.
\end{itemize}

\paragraph{Model Selection.}
To ensure optimal performance and fair comparison, we employ different checkpoint selection strategies based on the training objective.
For all standard regression models (including baselines and the pretrained base model), we select the checkpoint that achieves the lowest validation loss. 
Conversely, for \ours, we select the checkpoint that yields the highest mean rewards on the validation set.

\paragraph{Inference and Rollout Settings.} 
During the reinforcement learning phase (rollout), we employ stochastic sampling to encourage exploration.
\begin{itemize}
    \item \textbf{Rollout:} We set the sampling temperature to $1.0$. The number of samples generated per input is set to $16$ for Tabular Regression and $4$ for Code-to-Metric Regression.
    \item \textbf{Evaluation:} For final inference on the test set, we maintain the temperature at $1.0$ and aggregate predictions using the median of the generated candidates. The sampling budget is increased to $128$ samples for Tabular Regression and $64$ samples for Code-to-Metric Regression to ensure robust estimation.
\end{itemize}

\section{Additional Experiment}
\label{Appdix:extra-exp}
\subsection{Ablation on Tokenizer}
\label{Appdix:tokenizer}
\begin{table}[ht!]
\centering
\caption{
    Ablation study on different output tokenization schemes comparing R${}^2$ and Rank Correlation. The results are reported as the average over 5 random seeds across 100 TALENT regression tasks. The best results are \textbf{bolded}. 
    Rows shaded in \colorbox{gray!15}{gray} highlight the ReMax results for direct comparison against CE.
}
\label{tab:tokenization}

\newcommand{\cg}{\cellcolor{gray!10}}

\begin{tabular}{lll cccc}
\toprule
\multirow{2.5}{*}{\textbf{Tokenization}} & \multirow{2.5}{*}{\textbf{Metric}} & \multirow{2.5}{*}{\textbf{Method}} & \multicolumn{4}{c}{\textbf{Aggregation Strategy}} \\
\cmidrule(lr){4-7}
& & & \textbf{Median} & \textbf{Median + Filter} & \textbf{Mean} & \textbf{Mean + Filter} \\
\midrule

\multirow{4}{*}{Norm.} 
& \multirow{2}{*}{R$^2$} 
  & Base 
  & 0.6124 & / & 0.6368 & / \\
& & \cg \ours-ReMax 
  & \cg \textbf{0.6459} & \cg / & \cg \textbf{0.6508} & \cg / \\
\cmidrule(lr){2-7} 

& \multirow{2}{*}{Rank Corr.} 
  & Base 
  & 0.7705 & / & 0.7670 & / \\
& & \cg \ours-ReMax 
  & \cg \textbf{0.7785} & \cg / & \cg \textbf{0.7728} & \cg / \\

\midrule

\multirow{4}{*}{P10} 
& \multirow{2}{*}{R$^2$} 
  & Base 
  & $\text{-2.46} \times \text{10}^{\text{9}}$ & 0.5874 & $\textbf{-6.75} \times \textbf{10}^{\textbf{24}}$ & 0.6102 \\
& & \cg \ours-ReMax 
  & \cg \textbf{0.6057} & \cg \textbf{0.6123} & \cg $\text{-4.20} \times \text{10}^{\text{25}}$ & \cg \textbf{0.6251} \\
\cmidrule(lr){2-7}

& \multirow{2}{*}{Rank Corr.} 
  & Base 
  & 0.7630 & 0.7630 & 0.7161 & 0.7605 \\
& & \cg \ours-ReMax 
  & \cg \textbf{0.7862} & \cg \textbf{0.7862} & \cg \textbf{0.7675} & \cg \textbf{0.7692} \\

\midrule

\multirow{4}{*}{IEEE} 
& \multirow{2}{*}{R$^2$} 
  & Base 
  & $\text{-2.95} \times \text{10}^{\text{17}}$ & 0.5947 & $\text{-2.76} \times \text{10}^{\text{17}}$ & 0.6199 \\
& & \cg \ours-ReMax 
  & \cg $\textbf{-1.49} \times \textbf{10}^{\textbf{17}}$ & \cg \textbf{0.6179} & \cg $\textbf{-1.52} \times \textbf{10}^{\textbf{17}}$ & \cg \textbf{0.6307} \\
\cmidrule(lr){2-7}

& \multirow{2}{*}{Rank Corr.} 
  & Base 
  & 0.7652 & 0.7652 & 0.7389 & 0.7619 \\
& & \cg \ours-ReMax 
  & \cg \textbf{0.7769} & \cg \textbf{0.7769} & \cg \textbf{0.7604} & \cg \textbf{0.7701} \\

\bottomrule
\end{tabular}
\end{table}

We additionally evaluate the performance of \ours-ReMax across different output tokenization schemes, i.e., P10~\citep{P10} and IEEE floating-point representations~\citep{IEEE}.
Given that P10 and IEEE tokenization could yield outlier predictions due to the model's hallucinations~\citep{omnipred, decoding_regression}, we also include an outlier filtering strategy for the generated candidates.
From the R$^2$ and rank correlation results in Table~\ref{tab:tokenization}, we find that \ours, based on ReMax, consistently outperforms the base model under different output tokenizations, except for the mean R$^2$ on P10, showing the robustness of \ours-ReMax.
Besides, it is worth mentioned that tokenization with unlimited output range, e.g., P10 or IEEE, is easier to produce outlier values, resulting in poor R$^2$. 
However, such tokenization schemes remain higher rank correlation, implicitly capturing the relationship between numbers. 
We also observe that \ours-ReMax mitigates outliers in most cases, even obtaining positive median R$^2$ for P10, but it cannot eliminate the hallucinations.
Reducing the hallucinations for unbounded tokenization is still a crucial future work for decoding-based regression~\citep{omnipred, decoding_regression}.

\subsection{Robustness of \ours-ReMax Across Different Tokenizer Bases}
\label{Appendix:robust-ReMax}
\begin{figure}[t!]
    \centering
    \includegraphics[width=\linewidth]{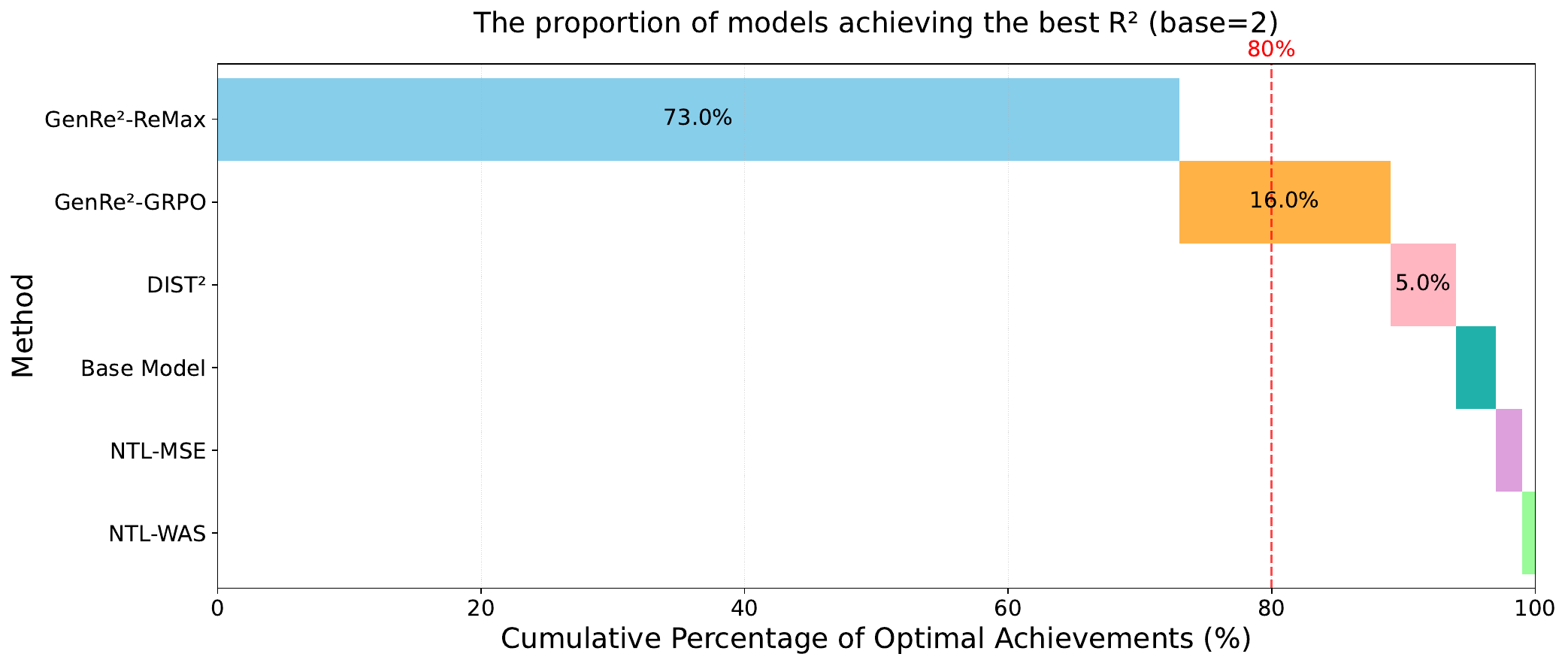}
    \caption{The proportion of models achieving the best R${}^2$. The length of each bar represents the proportion of the 100 datasets (in which a given method achieved the highest R${}^2$) on the TALENT benchmark. Note that all models utilized a normalized tokenizer with base=2.}
    \label{fig:digit-base2}
\end{figure}

\begin{figure}[t!]
    \centering
    \includegraphics[width=\linewidth]{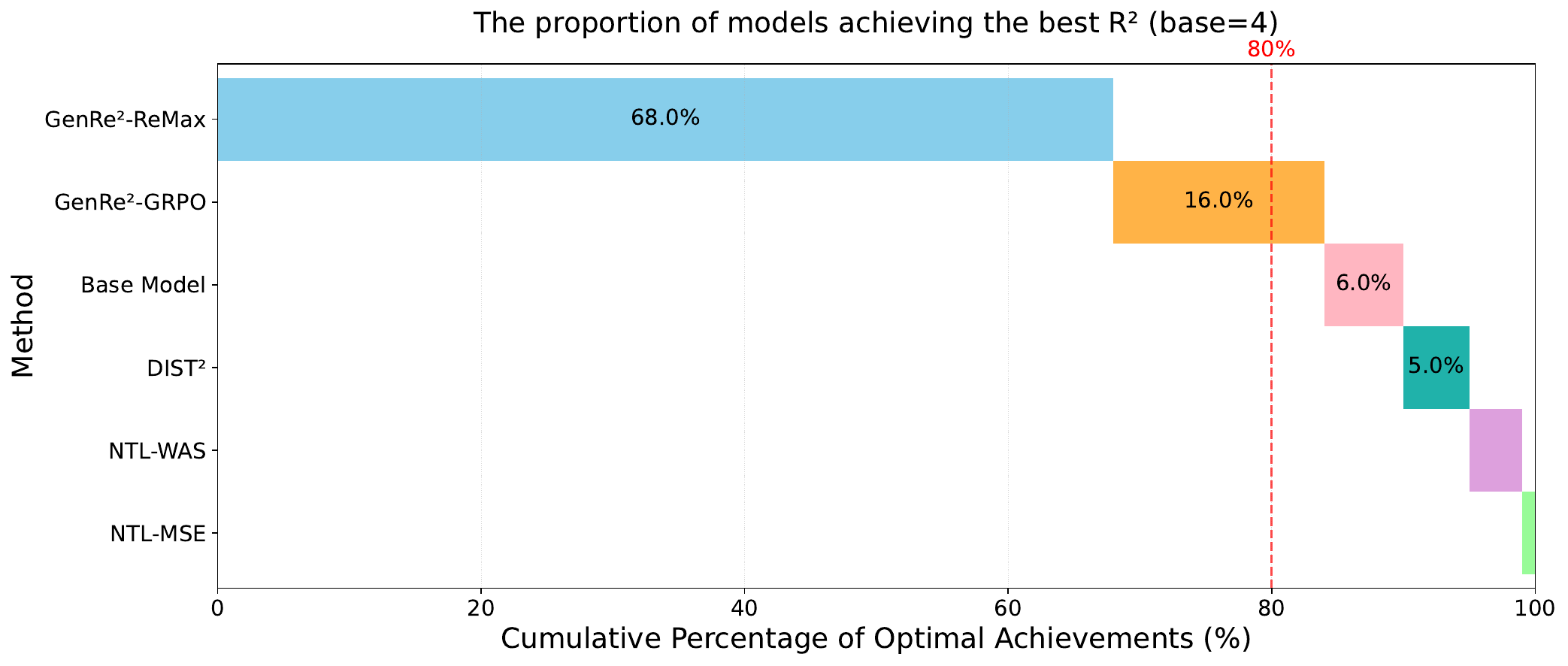}
    \caption{The proportion of models achieving the best R${}^2$. The length of each bar represents the proportion of the 100 datasets (in which a given method achieved the highest R${}^2$) on the TALENT benchmark. Note that all models utilized a normalized tokenizer with base=4.}
    \label{fig:digit-base4}
\end{figure}

\begin{figure}[t!]
    \centering
    \includegraphics[width=\linewidth]{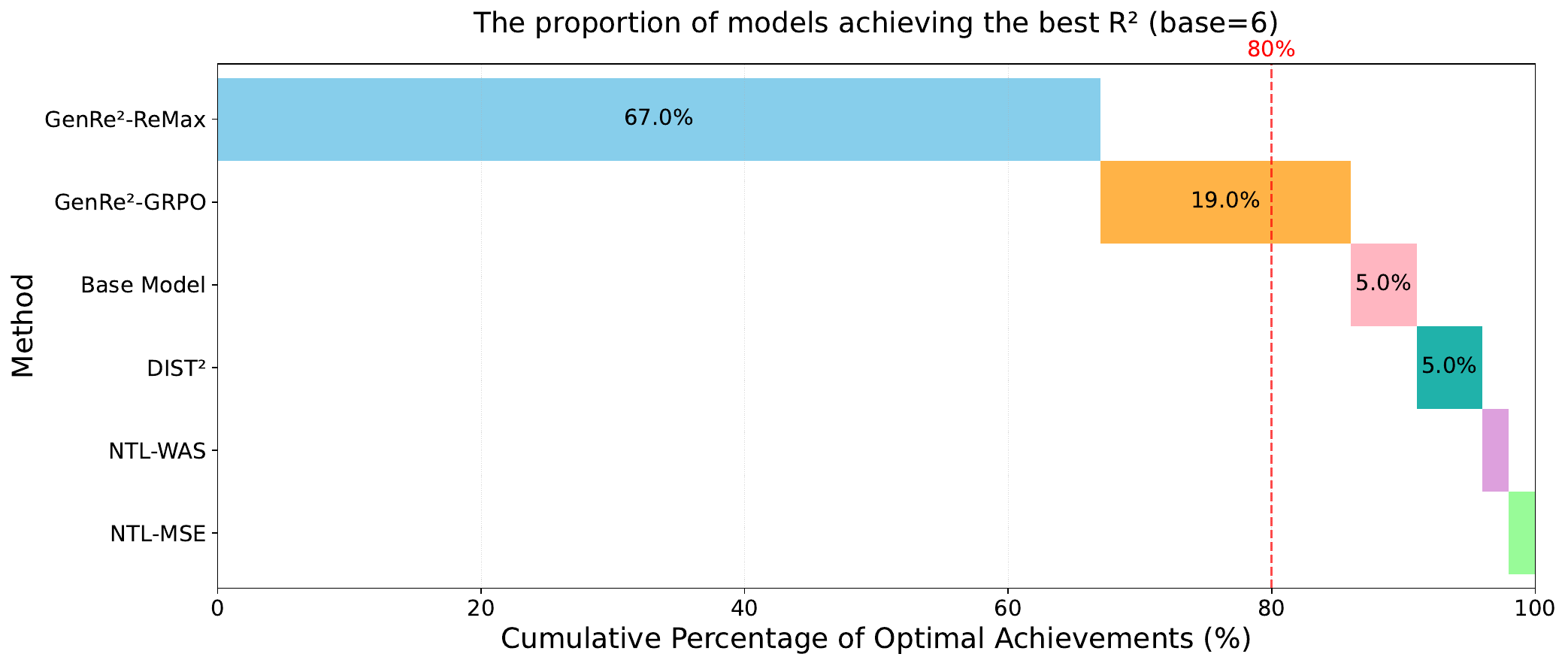}
    \caption{The proportion of models achieving the best R${}^2$. The length of each bar represents the proportion of the 100 datasets (in which a given method achieved the highest R${}^2$) on the TALENT benchmark. Note that all models utilized a normalized tokenizer with base=6.}
    \label{fig:digit-base6}
\end{figure}

\begin{figure}[t!]
    \centering
    \includegraphics[width=\linewidth]{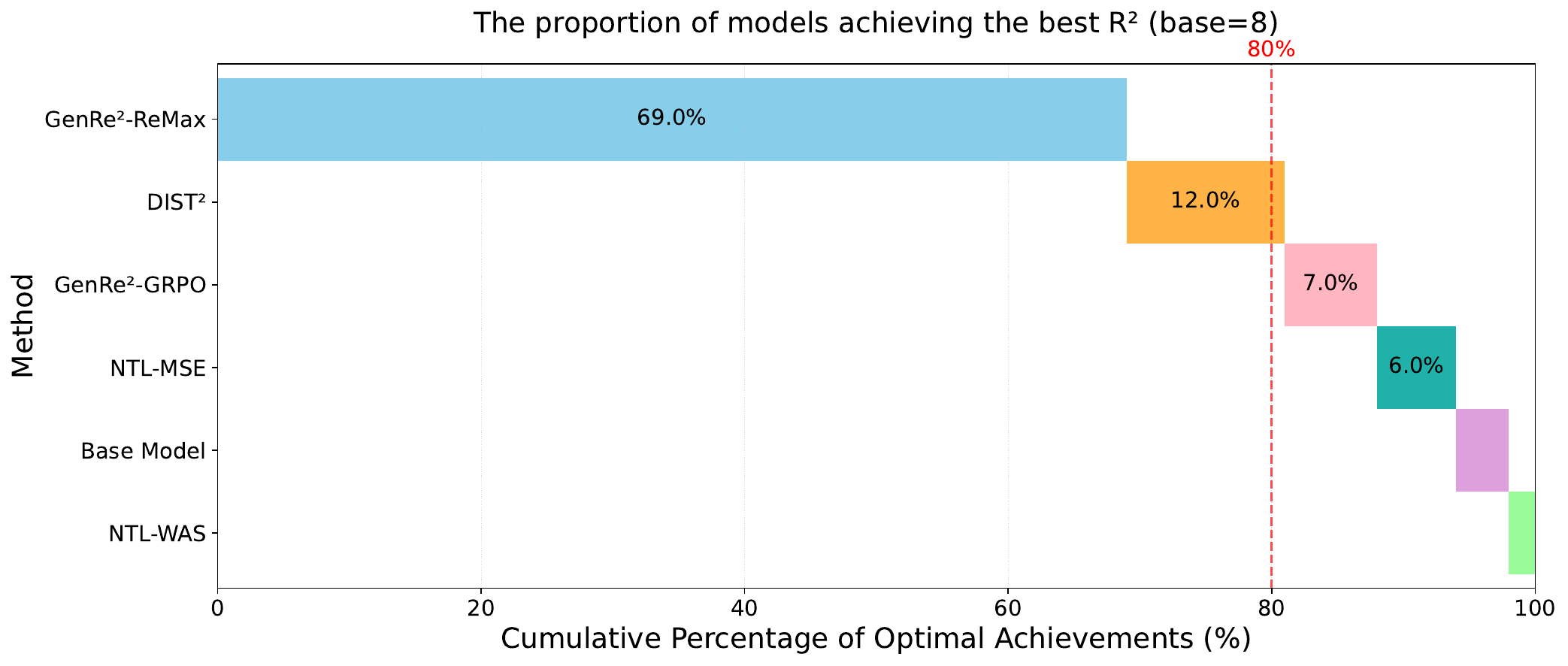}
    \caption{The proportion of models achieving the best R${}^2$. The length of each bar represents the proportion of the 100 datasets (in which a given method achieved the highest R${}^2$) on the TALENT benchmark. Note that all models utilized a normalized tokenizer with base=8.}
    \label{fig:digit-base8}
\end{figure}

\begin{figure}[t!]
    \centering
    \includegraphics[width=\linewidth]{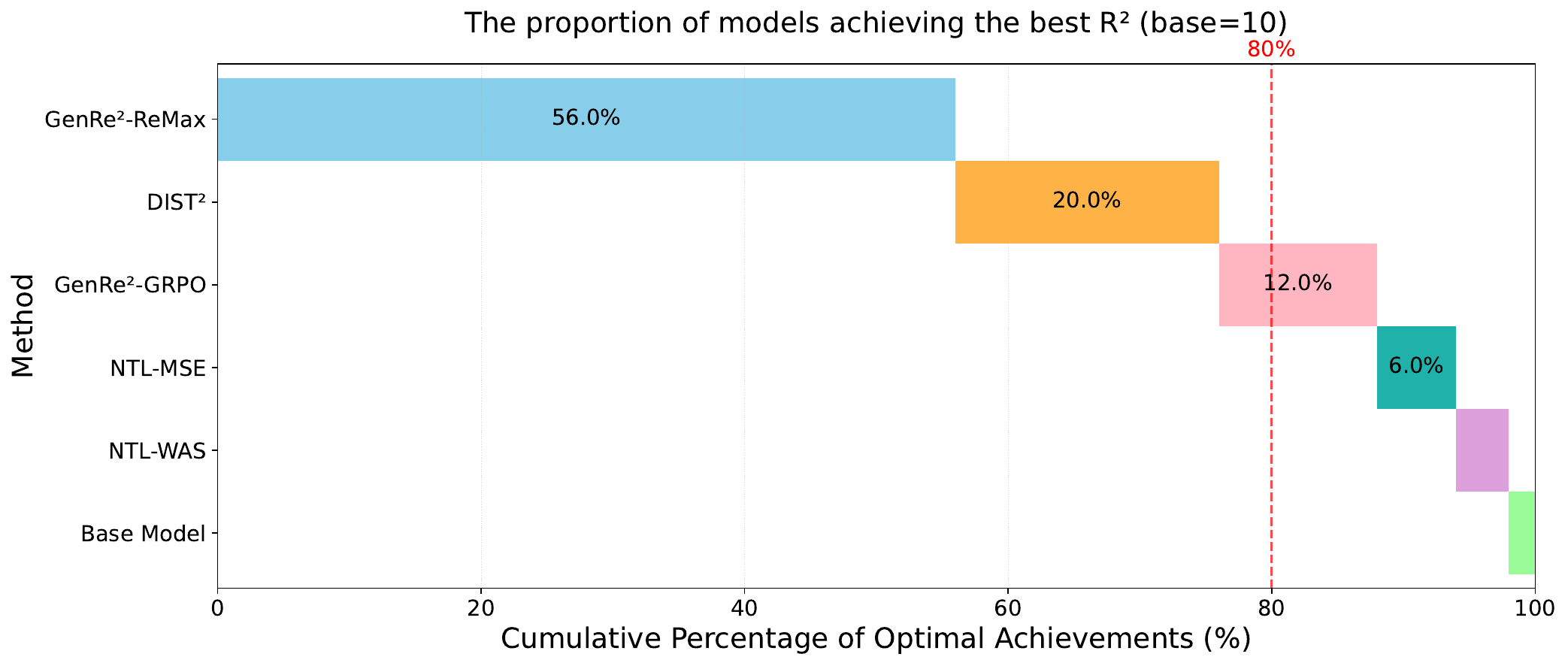}
    \caption{The proportion of models achieving the best R${}^2$. The length of each bar represents the proportion of the 100 datasets (in which a given method achieved the highest R${}^2$) on the TALENT benchmark. Note that all models utilized a normalized tokenizer with base=10.}
    \label{fig:digit-base10}
\end{figure}

We further investigate the impact of the tokenizer's base parameter on model ranking within the TALENT benchmark. As shown in Figures~\ref{fig:digit-base2} to~\ref{fig:digit-base10}, \ours-ReMax consistently achieves the highest R${}^2$ score on the majority of the 100 datasets, regardless of the base selected. 
Specifically, \ours-ReMax maintains a dominant position, securing the best performance across all configurations.
This analysis confirms that \ours-ReMax performs consistently well across different tokenizer settings, showing that it does not require specific tuning of the base parameter to achieve best results.


\subsection{Visualization of Target Normalization for Code Metric Regression}
\label{appendix:code-normalization}
\begin{figure}[!t]
    \centering
    \includegraphics[width=0.75\linewidth]{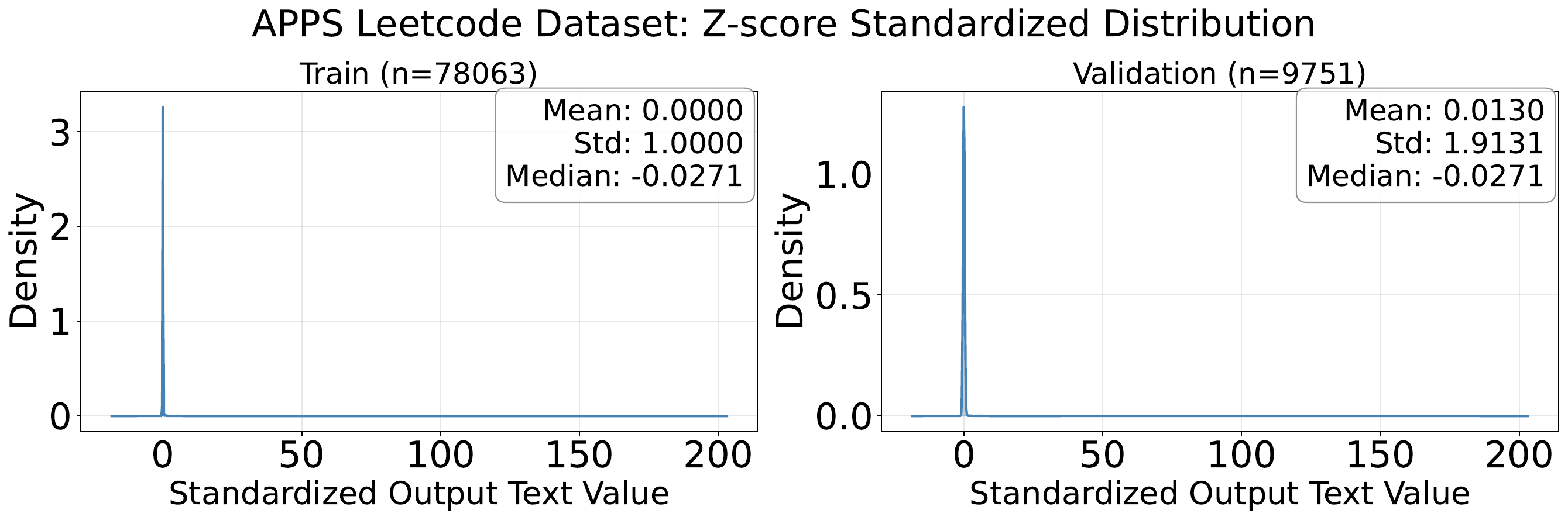}\vspace{1em}\\
    \includegraphics[width=0.75\linewidth]{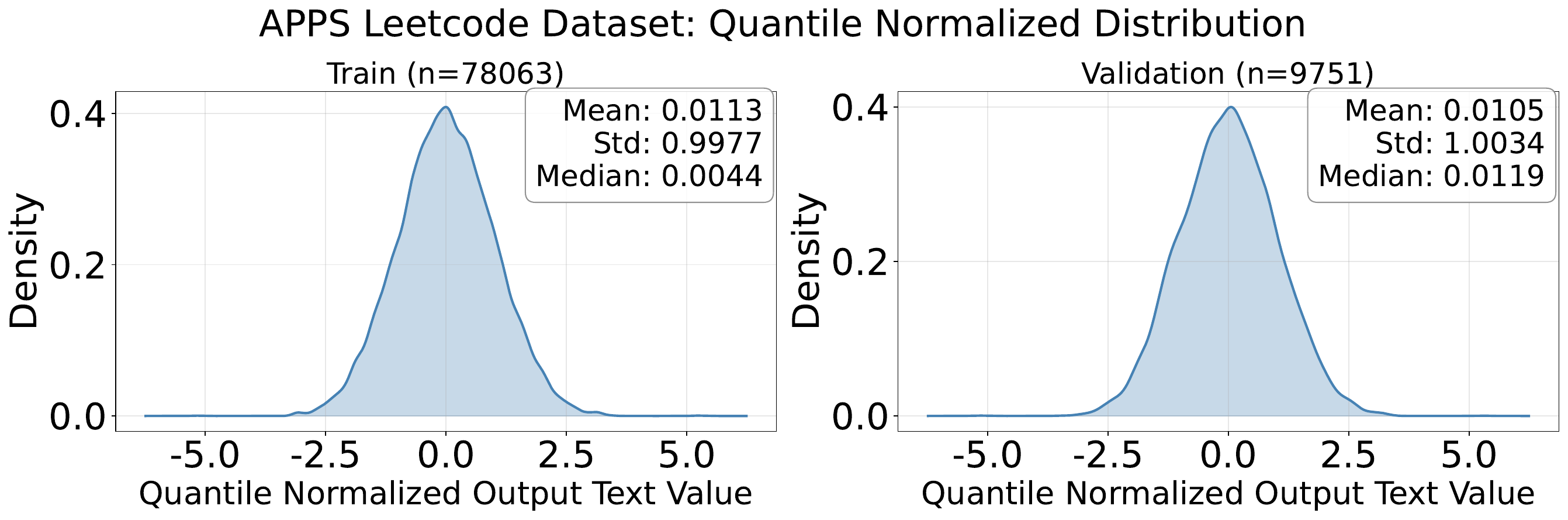} 
    \caption{Comparison of target value distributions on the APPS Leetcode Dataset across training and validation sets. \textbf{Top row}: Z-score standardization results in distributions with heavy tails and extreme outliers in both the training (left) and validation (right) splits. \textbf{Bottom row}: In contrast, quantile normalization effectively transforms the target values into a well-formed standard normal distribution consistently across both subsets.}
    \label{fig:quantile}
    \vspace{-1em}
\end{figure}

\begin{figure}[!t]
    \centering
    \includegraphics[width=0.75\linewidth]{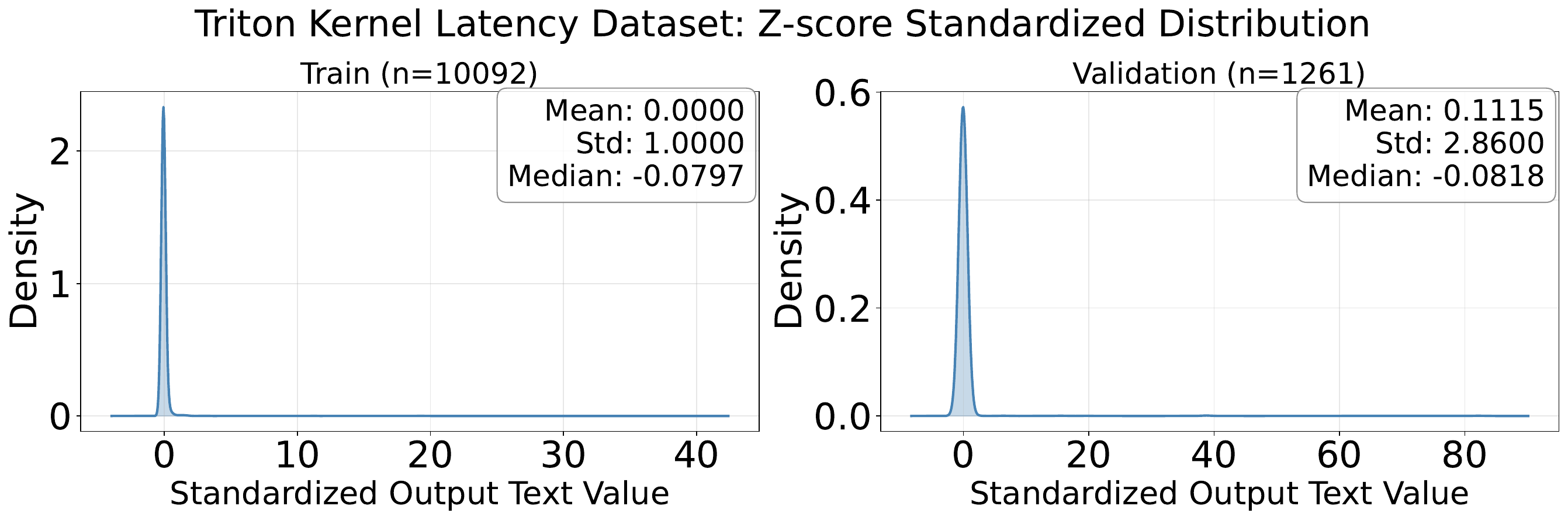}\vspace{1em}\\
    \includegraphics[width=0.75\linewidth]{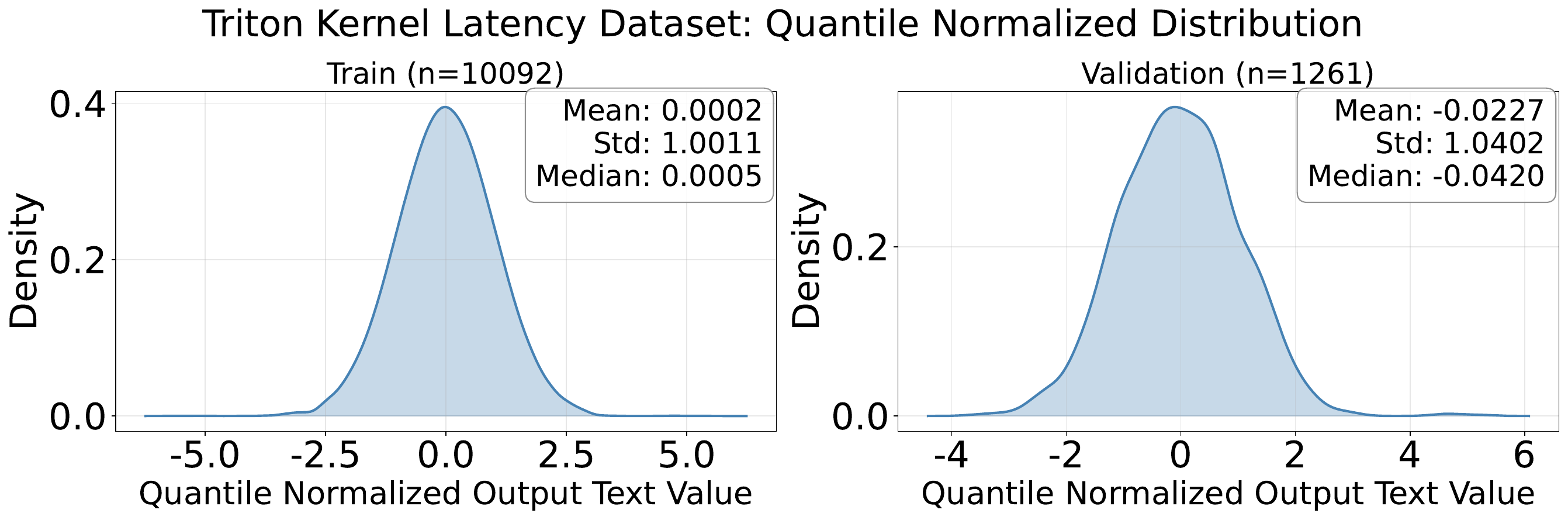} 
    \caption{Comparison of target value distributions on the Triton Kernel Latency Dataset across training and validation sets. \textbf{Top row}: Z-score standardization results in distributions with heavy tails and extreme outliers in both the training (left) and validation (right) splits. \textbf{Bottom row}: In contrast, quantile normalization effectively transforms the target values into a well-formed standard normal distribution consistently across both subsets.}
    \label{fig:quantile-kbss}
    \vspace{-1em}
\end{figure}

In this subsection, we visualize the target distribution under different normalization strategies mentioned in Section~\ref{sec:code-settings}.
As shown in Figures~\ref{fig:quantile} and~\ref{fig:quantile-kbss}, we can observe that both on the APPS Leetcode and the Triton Kernel Latency dataset, the z-score standardization exhibits sharp distribution and is prone to outliers, while the quantile normalization based on Gaussian delivers a smooth one.

\end{document}